\documentclass[11pt,english,authoryear]{elsarticle}
\usepackage[latin9]{inputenc}
\usepackage[a4paper]{geometry}
\usepackage[export]{adjustbox}
\usepackage[T1]{fontenc}
\usepackage{microtype}
\usepackage{float}
\usepackage{babel}
\usepackage{amsthm}
\usepackage{amsmath}
\usepackage{amssymb}
\usepackage{stmaryrd}
\usepackage{stackrel}
\usepackage{graphicx}
\usepackage{setspace}
\usepackage{accents}
\usepackage{booktabs,dcolumn}
\usepackage{caption}
\usepackage{subfig}
\usepackage{tikz}

\makeatletter
\def\ps@pprintTitle{%
  \let\@oddhead\@empty
  \let\@evenhead\@empty
  \let\@oddfoot\@empty
  \let\@evenfoot\@oddfoot
}
\makeatother

\floatstyle{ruled}
\newfloat{algorithm}{tbp}{loa}
\providecommand{\algorithmname}{Algorithm}
\floatname{algorithm}{\protect\algorithmname}

\theoremstyle{definition}
\newtheorem{defn}{\protect\definitionname}\theoremstyle{plain}
\newtheorem{lem}{\protect\lemmaname}\theoremstyle{plain}
\newtheorem{thm}{\protect\theoremname}\theoremstyle{plain}
\newtheorem{cor}{\protect\corollaryname}\theoremstyle{plain}
\newtheorem{prop}{\protect\propositionname}
\newtheorem*{rem*}{\protect\remarkname}

\providecommand{\definitionname}{Definition}
\providecommand{\lemmaname}{Lemma}
\providecommand{\theoremname}{Theorem}
\providecommand{\corollaryname}{Corollary}
\providecommand{\propositionname}{Proposition}
\providecommand{\remarkname}{Remark}

\DeclareMathOperator*{\argmin}{argmin}
\newcolumntype{.}{D{.}{.}{-1}}

\newcommand\thickbar[1]{\accentset{\rule{.4em}{.8pt}}{#1}}
\begin{document}
\title{Model selection consistency from the perspective of generalization ability and VC theory with an application to Lasso\tnoteref{tstar}}
\tnotetext[tstar]{The authors would like to thank Mike Bain, Colin Cameron, Peter Hall and Tsui Shengshang for valuable comments on an earlier draft. We would also like to acknowledge participants at the The 12\textsuperscript{th} International Symposium on Econometric Theory and Applications and The 26\textsuperscript{th} New Zealand Econometric Study Group and seminar participants at Utah, UNSW, and University of Melbourne for useful questions and comments. Fisher would like to acknowledge the financial support of the Australian Research Council grant DP0663477.}

\author{Ning Xu\corref{cor1}}
\ead{n.xu@sydney.edu.au}
\address{School of Economics, University of Sydney}

\author{Jian Hong\corref{cor2}}
\ead{jian.hong@sydney.edu.au}
\address{School of Economics, University of Sydney}

\author{Timothy C.G. Fisher\corref{cor3}}
\ead{tim.fisher@sydney.edu.au}
\address{School of Economics, University of Sydney}

\begin{abstract}
Model selection is difficult to analyse yet theoretically and empirically important, especially for high-dimensional data analysis. Recently the least absolute shrinkage and selection operator (Lasso) has been applied in the statistical and econometric literature. Consistency of Lasso has been established under various conditions, some of which are difficult to verify in practice. In this paper, we study model selection from the perspective of generalization ability, under the framework of structural risk minimization (SRM) and Vapnik-Chervonenkis (VC) theory. The approach emphasizes the balance between the in-sample and out-of-sample fit, which can be achieved by using cross-validation to select a penalty on model complexity.  We show that an exact relationship exists between the generalization ability of a model and model selection consistency. By implementing SRM and the VC inequality, we show that Lasso is $\mathcal{L}_2$-consistent for model selection under assumptions similar to those imposed on OLS. Furthermore, we derive a probabilistic bound for the distance between the penalized extremum estimator and the extremum estimator without penalty, which is dominated by overfitting. We also propose a new measurement of overfitting, $GR^2$, based on generalization ability, that converges to zero if model selection is consistent. Using simulations, we demonstrate that the proposed CV-Lasso algorithm performs well in terms of model selection and overfitting control.

\end{abstract}

\maketitle

\renewcommand{\baselinestretch}{1.2}

\vspace*{\fill}
\begin{center}
\Large

Model selection consistency from the perspective of generalization \\
ability and VC theory with an application to Lasso

\thispagestyle{empty}

\vspace{\fill}
\large
\today
\end{center}

\vspace{\fill}

\newpage
\setcounter{page}{1}
\section{Introduction}

Model selection is vital in econometric analysis for valid inference and accurate prediction. Moreover, given the increasing prevalence of high-dimensional data analysis in economics, model selection is coming to the forefront of statistical inference. With high-dimensional data, the curse of dimensionality  \citep{bellman57} becomes a concern. In econometrics, the curse of dimensionality refers to the difficulty of fitting a model when a large number of possible predictors ($p$) are available. When the dimension is high relative to the given sample size $n$, the effective sample size ($n/p$ or $n/\log(p)$) is relatively small, making it harder to sample the population space sufficiently. With a larger $p$, the model to be estimated becomes more complex as well. A model may perfectly fit the data when $p = n$, which is an example of the well-known overfitting problem. Estimation may also be affected by dimensionality in other ways. Estimation involving a matrix inverse, numerical integrals, or grid search may be difficult to implement with high-dimensional data. The convergence rate of non-parametric estimators is lower with a higher $p$. Problems due to measurement errors and missing values in estimation become worse with high-dimensional data as well. In this paper, we focus on linear model selection which reduces to variable selection and dimension reduction. However, the analysis covers some non-parametric models such as series regression and also provides an approximation to non-linear models in general---see \citet{belloni2011high}.

Model selection typically involves using a score function that depends on the data \citep{heckerman95}, as with the Akaike information criterion \citep{akaike73}, the Bayesian information criterion \citep{schwarz78}, cross-validation methods \citep{stone74,stone77}, and mutual information scores among variables (see \citet{friedman97} and \citet{friedman00}). \citet{shao97} proves that various types of information criterion (IC) and cross-validation are consistent in model selection. However, the optimization-based search algorithms that are often used to implement these methods are not without drawbacks. First, they tend to select more variables than necessary and, as illustrated by \citet{breiman95}, they are sensitive to small changes in the data. Second, especially with high-dimensional data, combinatorial search algorithms may be computationally challenging to implement.\footnote{As \citet{chickering04} points out, the best subset selection method is unable to deal with a large number of variables, heuristically 30 at most.}

As an alternative to conventional model selection methods, the least absolute shrinkage and selection operator (Lasso) is introduced by \citet{tibshirani96}. Consider the linear regression model
\[
Y = X\beta + u
\]
where $Y\in\mbox{Matrix}(n\times1,\mathbb{R})$ is a vector of response variables, $X\in\mbox{Matrix}(n\times p,\mathbb{R})$ is a matrix of covariates and $u\in\mbox{Matrix}(n\times1,\mathbb{R})$ is a vector of i.i.d.\ random errors. We are interested in estimating the parameter vector $\beta\in R^{p}$, which may be sparse in the sense that many of its elements are zero. The Lagrangian of the penalized least squares model may be written
\begin{equation}
\min_{b_\lambda}
\frac{1}{n}
\left(\left\Vert Y-Xb_{\lambda}\right\Vert _{2}\right)^{2}+\lambda\Vert b_{\lambda}\Vert_{\gamma}
\label{lasso}
\end{equation}
where $\Vert\cdot\Vert_{\gamma}$ is the $\mathcal{L}_{\gamma}$ norm and $\lambda\geqslant0$ is the penalty or tuning parameter. The estimator $b_{\lambda}$ is the solution to the constrained minimization problem. Note that if $\lambda=0$, the usual OLS estimator is obtained. Lasso corresponds to the case with $\gamma=1$. When $\gamma=2$, we have the familiar ridge estimator \citep{tikhonov1963}, which typically is not used for model selection. As a generalization of the ridge estimator, \citet{frankfriedman93} propose the bridge estimator for any $\gamma>0$. \citet{fu98} provides a comparison of these estimators in a simulation study.

Lasso may be thought of as a `shrinkage estimator'. \citet{james1961} prove that, on average, the shrinkage estimator dominates the OLS estimator in terms of mean squared error (MSE).\footnote{The ridge estimator  is an early example of a shrinkage estimator. The shrinkage estimator is also called the James-Stein estimator.} A shrinkage estimator restricts the norm of the estimated parameter vector to be less than or equal to a constant. By restricting $\left\Vert b_{\lambda}\right\Vert _{1}$ to be smaller than a constant, Lasso shrinks some $b_{i}$ to zero, effectively dropping the corresponding $X_{i}$ from the model. Surprisingly, a constrained estimator like Lasso may outperform an unconstrained estimator like OLS in terms of the bias-variance trade-off. From (\ref{lasso}) it is clear that Lasso will produce a different model for each value of the penalty parameter $\lambda$. In general, a higher value of $\lambda$ corresponds to a higher penalty and a smaller number of $X_i$. Thus the complexity of the model can be controlled by the value of $\lambda$. We use an algorithm where $\lambda$ is chosen by cross-validation, which we call the CV-Lasso algorithm.\footnote{See the pseudo code in Appendix~2. Alternatively, $\lambda$ can be chosen using the AIC or the BIC. Our simulations show that CV-Lasso slightly outperforms the BIC-Lasso, which in turn outperforms the AIC-Lasso.} In economics, we often observe only one sample: cross-validation divides the sample into training and test sets. The parameters of interest are estimated using the training set with a given value of the penalty parameter. The estimated model is then applied to the test set to calculate the associated loss. The selected model is based on the $\lambda$ with the lowest loss. \citet{varian14} advocates that cross-validation should be used much more in economics, particularly when working with large datasets, because it may provide a more realistic measure of prediction performance than measures commonly used in economics such as $R^2$.

A range of properties have been established for Lasso-type estimators. The parameter estimation and model selection consistency of Lasso are established for fixed $p$ by \citet{knightfu00}. \citet{meinshausenbuhlmann06} show that Lasso is consistent in the Gaussian scenario even when $p > n$. \citet{zhaoyu06} establish probabilistic consistency for both fixed $p$ and large $p$ problems. They find that Lasso selects exactly the set of nonzero regression coefficients under the `irrepresentable condition,' which may be hard to verify in practice. \citet{zhanghuang08} study the bias in Lasso and derive its consistency (convergence) rate.

Lasso is becoming increasingly popular in econometrics. As argued by \citet{varian14}, econometrics may require a different set of tools for manipulating and analyzing big data sets. Many tools from statistical learning can be adapted for econometric analysis; Lasso is one such tool. Recent applications of Lasso in economics include \citet{baing08}, \citet{demolall08}, \citet{pistoresiall11}, \citet{schneiderwagner12}, \citet{kimswanson14}, and \citet{manzan15}. \citet{belloniall12} propose using Lasso to select instruments while the parameters of interest are estimated by conventional procedures. \citet{caner09} proposes a Lasso-type GMM estimator and derives its asymptotic properties for the case where $0<\gamma<1$. \citet{chatterjeeall15} study the oracle property of the residual empirical process of the adaptive Lasso. \citet{kockcallot15} study the properties of Lasso and adaptive Lasso for a stationary VAR model with Gaussian errors. \citet{chengliao15} use Lasso to select moments where the penalty term depends on a preliminary consistent estimator that accounts for the strength and
validity of the moments.\footnote{The last three papers are in a recent {\em Journal of Econometrics} special issue on high-dimensional data problems in econometrics.}

In this paper, we study model selection from the perspective of generalization ability, the ability of a selected model to predict outcomes in new samples from the same population. Generalization ability is important for prediction purposes or for studying the effect of a new policy. The perspective is based on Vapnik-Chervonenkis (VC) theory \citep{vc71b}, a fundamental theory in statistical learning. In VC theory, an estimator (or algorithm) with good generalization ability will perform well with `in-sample' data and `out-of-sample' data. The consistency of model selection can be established under the structural risk minimization (SRM) framework, one of the main principles in VC theory. According to SRM, there are essentially two reasons why a model selected from one sample may not fit another sample well: the two samples may have different sampling errors, or the complexity of the model selected from the original sample may have been set inappropriately. To improve the generalization ability of the model estimated from a sample, SRM requires minimizing the error, known as the the `generalization error' (GE), when the estimated model is applied to another sample. The balance between in-sample and out-of-sample fitting is described by the `VC inequality'. We adapt and generalize the VC inequality (in Lemmas~1 and~2) for extreme estimators and establish a model-free and distribution-free probabilistic bound for the generalization error (in Theorem~1). We also propose a measurement based on generalization ability, $GR^{2}$, to summarize the in-sample and out-of-sample goodness-of-fit.

Using SRM, we then establish the consistency of Lasso-type model selection. For the $n \ge p$ case, the assumptions for consistency are similar to (and actually weaker than) those usually imposed on OLS, while for the $n < p$ case an additional assumption on sparse eigenvalues of the $X^T X$ matrix is required. Given a sample, SRM can be implemented in Lasso by selecting $\lambda$, which is equivalent to controlling the complexity of the model. We show that, under certain conditions, the true DGP uniquely offers the minimum generalization error in the population (Proposition~1). Hence, we show that the true DGP will be selected by Lasso given $\lambda$ (Proposition~2). We then show (Theorems~2, 3, and~4) that the VC inequality and minimization of the empirical GE guarantees not only that Lasso is consistent in model selection, but also that Lasso offers a better out-of-sample fit than extremum estimators. We derive a probabilistic bound for the distance between the penalized extremum estimator and the extremum estimator without penalty, which is dominated by overfitting. We have a detailed discussion on how the choice of $\lambda$ affects model selection.

Our proof strategy highlights the connection between asymptotic performance and generalization ability. Instead of restricting attention to a single sample, we consider both in-sample and out-of-sample fit. Then we transform and reformulate the consistency problem into the GE space. We show that empirical GE minimization not only controls overfitting and improves the finite-sample performance, but also helps us to find the true model asymptotically. In addition, our method has the potential to extend the consistency results in \citet{knightfu00}, \citet{zhaoyu06}, \citet{candestao07} and \citet{meinshausenyu09} to functional regression. Furthermore, our work also sheds light on the applicability of general model
selection based on VC theory, offering insights into the bias-variance trade-off from the perspective of generalization ability.

The paper is organized as follows. We first discuss the relation between generalization ability and model selection consistency in section~2. In section~3, we prove that Lasso is $\mathcal{L}_{2}$-consistent in model selection under the proposed conditions. In section~4, we use simulations to demonstrate the ability of Lasso to select models and control for overfitting. Section~5 concludes with a brief discussion of our results. Proofs are contained in Appendix~1, pseudo-code for the algorithms is in Appendix~2, and graphs of the simulations are in Appendix~3.

\section{Generalization ability, structural risk minimization and model selection }

\subsection{Generalization ability, and overfitting }

In econometrics, choosing the best approximation to data involves measuring a loss, $\mathrm{Loss}(y_{i},\widehat{m}(x_{i},b)),\,i=1,\ldots,n$, defined as a functional between the estimated value $\widehat{m}(x,b)$ and the true value $y$. The risk functional is defined as
\[
\mathcal{R}(b\vert X,Y)=\int\mathrm{Loss}(y,\widehat{m}(x,b))\mathrm{d}F(x,y)
\]
where $F(x,y)$ is the joint distribution of $(x,y)$. Without knowing the distribution $F(x,y)$ a priori, we define the empirical risk functional as follows
\[
\mathcal{R}_{n}(b\vert X,Y) = \frac{1}{n}\;\sum_{i=1}^n\;\mathrm{Loss}(y_{i},\widehat{m}(x_{i},b)).
\]
In the regression case, for example, the estimated value $\widehat{m}(x,b)=\hat{y}=X\hat{b}$ and $\mathcal{R}_{n}(b\vert X,Y)=\frac{1}{n}\sum_{i=1}^n(y_i - \hat{y}_i)^{2}$.

For regression models, the $R^{2}$ is often used to measure goodness-of-fit for in-sample data. We can rewrite $R^{2}$ as $1-\mathcal{R}_{n}(b\vert X,Y)/\mbox{TSS}$ where $\mbox{TSS}=(1/n) \sum_{i=1}^n (y-\bar{y})^{2}$. For high-dimensional data analysis, however, an estimated model with a high $R^{2}$ may have poor predictive power with out-of-sample data, a feature commonly referred to as `overfitting'. As a result, in-sample fit may not be a reliable indicator of the general usefulness of the model. Thus, \citet{vc71} propose the \textbf{generalization ability} (GA) of a model, a measure of its prediction performance with out-of-sample data.

Generalization ability can be measured by different criteria. In the case where $X$ and $Y$ are directly observed, generalization ability is a function of the difference between the actual $Y$ and the estimated $Y$ for out-of-sample data. In this paper, generalization ability is measured by the generalization error (GE).\footnote{In the statistical learning literature, GE is also referred to as the `test error' or `validation error'.} Generally speaking, GE can be defined in terms of empirical risk.
\begin{defn}
The $\mathcal{L}_{2}$ \textbf{training error} is defined as $\min_b\mathcal{R}_{n_t}(b|Y_{t},X_{t}) = \mathcal{R}_{n}(b_{train}|Y_{t},X_{t})$ where $b_{train}$ minimizes $\mathcal{R}_{n_t}(b|Y_{t},X_{t})$ and $(Y_{t},\,X_{t})$ refers to the data used for the estimation of $b$, also called the \textbf{training set}. The $\mathcal{L}_{2}$ \textbf{generalization error} is defined as $\mathcal{R}_{n_s}(b_{train}|Y_{s},X_{s})$ where $(Y_{s},\,X_{s})$ refers to data that is not used for the estimation of $b$, also called the \textbf{test set}.
\end{defn}
For linear regression, the estimator, training error, and generalization errors are, respectively as follows:
\begin{align*}
                               b_{train} &=\argmin_b\,\frac{1}{n_t}\,\Vert Y_{t}-X_{t}b\Vert_{2}^{2}\\
\mathcal{R}_{n_t}(b_{train}|Y_{t},X_{t}) &=\frac{1}{n_t}\,\Vert Y_{t}-X_{t}b_{train}\Vert_{2}^{2}\\
\mathcal{R}_{n_s}(b_{train}|Y_{s},X_{s}) &=\frac{1}{n_s}\,\Vert Y_{s}-X_{s}b_{train}\Vert_{2}^{2}
\end{align*}
where $n_t$ and $n_s$ are the sample sizes for the training set and the test set, respectively. Henceforth, $\min\{n_s,n_t\}$ is denoted by $\widetilde{n}$.

If we have multiple samples, it is straightforward to define some of them as test sets and others as training sets, use training sets for estimation and use test sets to validate the generalization ability of the model estimated from the training sets. This method is called `validation'. If we only collect one sample from the population, we can randomly partition it into two subsets: one as the training set and the one as the test set. However, in reality we may not have enough sample points for validation with such a partition. To put this another way, if the only sample we collect is not large enough and we partition it into training and test sets, we decrease the size of the training set and consequently affect the performance of the model we estimate from training sets. Hence, when we have only one sample and its size is not large enough to support such random partition, we need to switch to $K$-fold cross validation.

In more detail, cross validation implies randomly partitioning the full sample into $K$ folds.\footnote{Typically, $K=5,10,20,40\mbox{ or }N$.} We choose one fold as the test set, and designate the remaining $K-1$ folds as the training set. We then carry out extremum estimation on the training data and use the fitted model to record its GE on the test set. This process is repeated $K$ times, with each of the $K$ folds getting the chance to play the role of the test set, with the remaining $K-1$ folds used as the training set. In this way, we obtain $K$ different estimates of the GE for the fitted model. These $K$ estimates of the GE are averaged, giving the cross-validated GE.

By implementing cross validation, each data point is used in both the training and the test sets. Moreover, cross validation reduces the resampling error by running validation $K$ times over different training and test sets. Hence, intuitively, cross validation is more robust on resampling error and should perform at least as well as validation. In section~3, we study the generalization error of penalized extremum estimators in both the validation and cross validation cases and show the difference between them in detail.

We use the training error to measure \textit{in-sample fit} and the generalization error to measure \textit{out-of-sample fit}. The two errors illustrate why the generalization ability of a model is crucial to model selection. When an unnecessarily complicated model is imposed on the data, it will generally suffer from overfitting: the model will be too tailored for in-sample data, compromising its out-of-sample performance. To summarize the in-sample and out-of-sample goodness of fit, we propose the following empirical measure
\begin{equation}
GR^{2} =
\left(1 - \frac{\mathcal{R}_{n_s}(b_{train}|Y_{s},X_{s})}{\mathrm{TSS}\left(Y_{s}\right)}\right)
\times
\left(1 - \frac{\mathcal{R}_{n_t}(b_{train}|Y_{t},X_{t})}{\mathrm{TSS}\left(Y_{t}\right)}\right) \\
= R^2_s \times R^2_t
\label{grsq}
\end{equation}
where $R^2_s$ is the the $R^2$ for the test set, and $R^2_t$ is the $R^2$ for the training set. If $b_{train}$ is consistent, both $\mathcal{R}_{n_t}(b_{train}|Y_{t},X_{t})$ and $\mathcal{R}_{n_s}(b_{train}|Y_{s},X_{s})$ converge to the same limit in probability as $\widetilde{n}\rightarrow\infty$, implying that ${\lim}_{\widetilde{n}\rightarrow\infty} GR^{2}=1$.

Clearly $GR^2$ combines measures of the in-sample fit and the out-of-sample fit. Intuitively, there are four different possibilities for $GR^2$. A model that fits the training set and the test set well will have high $R^2_t$ and $R^2_s$ values and hence a high $GR^2$. When overfitting occurs, the $R^2_t$ will be relatively high and the $R^2_s$ will be low, reducing the $GR^2$. When underfitting occurs, both the $R^2_t$ and $R^2_s$ will be low, reducing the $GR^2$ further. It is also possible that the model estimated on the training set fits the test set better (the $R^2_s$ is high while the $R^2_t$ low). In the section~4 simulations we find that the $GR^2$ performs well as a measure of overfitting and underfitting.

\subsection{Structural risk minimization and model selection}

In econometrics, choosing the best model for data typically involves minimization of the training error $\mathcal{R}_{n}(b)$, which is also the SRM principle proposed by \citet{vc71,vc71b}. Essentially, the SRM principle states that: given the functional form $\widehat{m}$, the sampling error (that is, error due to the empirical distribution) $\Vert\mathcal{R}_{n}(b\vert X,Y)-\mathcal{R}(b\vert X,Y)\Vert$ converges to zero as the sample size increases. If $\widehat{m}(x,b)$ happens to be the correct functional form for the model, the SRM principle is equivalent to the consistency property in econometrics.

The relation between $\mathcal{R}_{n}(b)$ and $\mathcal{R}(b)$ is summarized by the VC inequality \citep{vc74b} as follows.
%
%
\begin{lem}
{\rm \citep{vc71}.} The following VC inequality holds with probability (or power) $1-\eta$, $\forall b$, $\forall n\in\mathbb{N}^{+}$,
\begin{equation}
\mathcal{R}(b|X,Y) \leqslant\frac{\mathcal{R}_{n_t}(b|X_{t},Y_{t})}{1-\sqrt{\epsilon}}
\label{lem1}
\end{equation}
or
\begin{equation}
\mathcal{R}(b|X,Y) \leqslant\mathcal{R}_{n_t}(b|X_{t},Y_{t})
+\frac{\sqrt{\epsilon}}{1-\sqrt{\epsilon}}\mathcal{R}_{n_t}(b|X_{t},Y_{t})
\label{lem1a}
\end{equation}
where $\mathcal{R}_{n_t}(b|X_t,Y_t)$ is the training error from the extremum estimator $b$, $\mathcal{R}(b|X,Y)$ is the expectation of the generalization error $\mathcal{R}_{n_s}(b|X_{s},Y_{s})$, $h$ is the VC dimension for $b$, and $\epsilon=(1/n_t)[h\ln(n_t/h)+h-\ln\left(\eta\right)]$.
\end{lem}
The VC dimension is a measure of the complexity of the model and reduces to $p$ for the case of generalized linear models.\footnote{In classification models, the VC dimension is different from $p$, see \citet{vc74b}.} As long as $h$ for the model is finite, the model will never result in an $R^{2} = 1$ or $GR^{2} = 1$ regardless of the sample. A detailed explanation of $h$ can be found in the proof of Theorem~1 in Appendix~1. As shown in Figure~1, the VC inequality provides an upper bound for the generalization error of $b$. When the effective sample size, defined as $n_t/h$, is large, $\epsilon$ is small, the second term on the RHS of (\ref{lem1a}) becomes small, the training error is close to the generalization error, and overfitting is inconsequential (or can be ignored). However, if the effective sample size $n_t/h$ is small (that is, the model is very complicated), the second term on the RHS of (\ref{lem1a}) becomes larger. In such situations a small training error does not guarantee a small generalization error and overfitting becomes more likely.

\begin{figure}
\centering
\includegraphics[width=0.45\paperwidth]{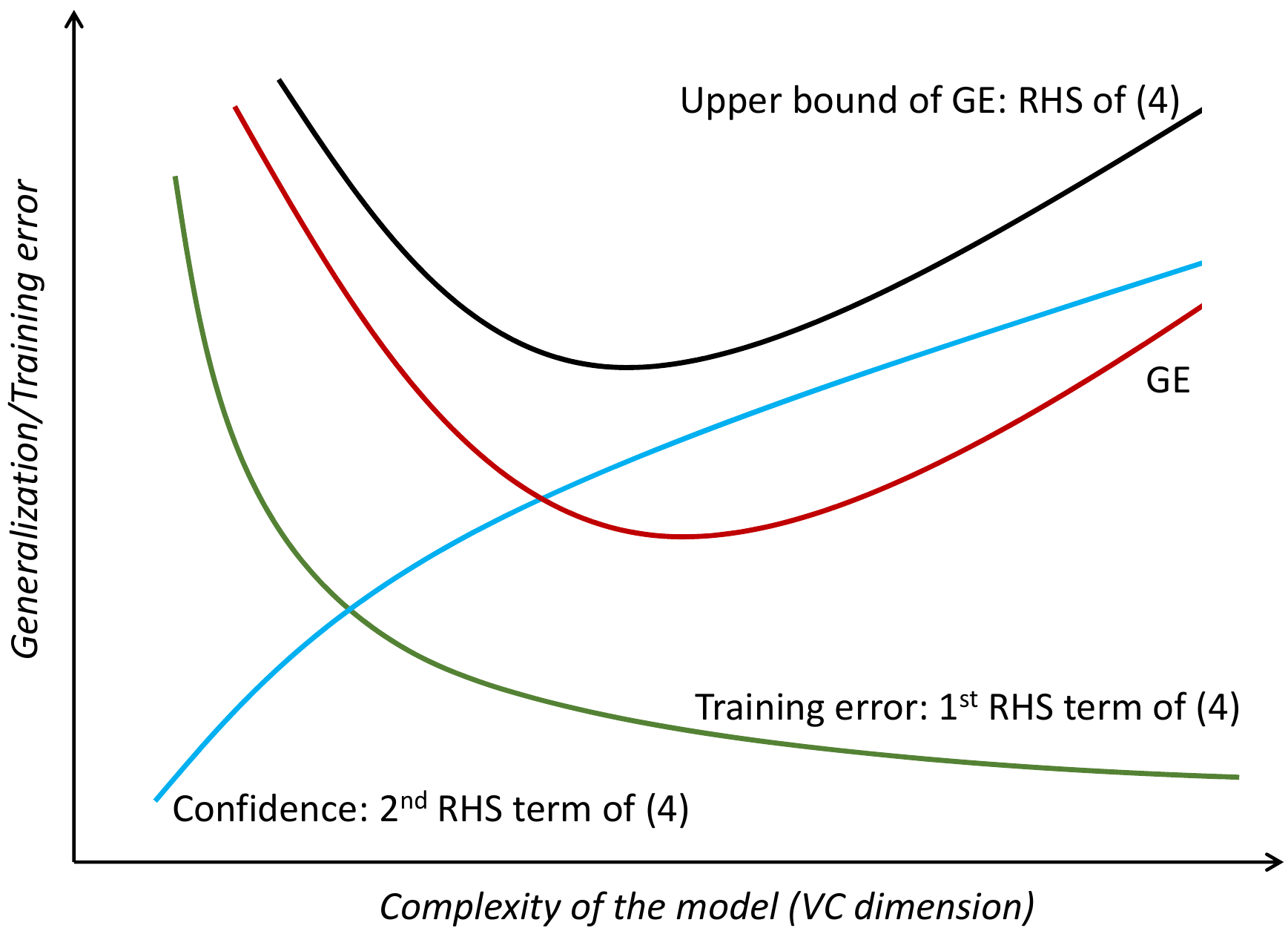}

\caption{The VC inequality and structural risk minimization}
\end{figure}

In the small $n_t/h$ case, reducing overfitting requires minimizing both terms on the RHS of (\ref{lem1a}). Since the second term in (\ref{lem1a}) depends on $h$, it follows that, instead of minimizing $\mathcal{R}_{n_t}$, it is necessary to minimize the upper bound of the GE. \citet{vc71} show that SRM guarantees that the minimal GE chosen by SRM converges to the minimum GE in the population at a given rate, as shown below in Theorem~1. Here we denote the model chosen by SRM as $b_{SRM}$ and $\Lambda$ as the space of alternative models.
%
%
\begin{lem}
{\rm \citep{vc71}.} SRM provides approximations for which the sequence of
$\mathcal{R}_{n_t}\left(b_{SRM}|X_{t},Y_{t}\right)$ converges to the smallest generalization error
\[ \mathcal{R}_{min} = \inf_{b\in\Lambda} \int\mathrm{Loss}(b\vert X,Y)\mathrm{d}F(x,y) \]
with asymptotic rate of convergence
\[ V(n_t) = r_{n_t} + \tau\sqrt{\frac{h\ln(n_t)}{n_t}}, \]
if
\[ \lim_{n_t\rightarrow\infty} \frac{\tau^{2}h\ln(n_t)}{n_t} = 0, \]
where $F(x,y)$ is the population distribution of $(X,Y)$, $\tau$ is a positive number such that, for $1<p$,
\[
\tau\geqslant
\sup_{b\in\Lambda} \frac{[\int\left(\mathrm{Loss}(b|X,Y)\right)^p}{\mathrm{d}F(x,y)]^{1/p}}
{\int\mathrm{Loss}(b|X,Y)\mathrm{d}F(x,y)},
\]
and
\[
r_{n_t}=\mathcal{R}_{n_t}(b_{SRM}|X_t,Y_t)-\inf_{b\in\Lambda}\int\mathrm{Loss}(b|X,Y)\mathrm{d}F(x,y).
\]
\end{lem}
VC dimension is crucial for SRM because it is used to construct the upper bound for the generalization error. SRM has been implemented to reduce overfitting in classification models for many years but it can be hard to implement in other models because it is difficult to calculate the VC dimension.\footnote{The VC dimension is known for only 3 types of models, including linear regression.} Researchers in statistics have ignored the upper bound of the generalization error and have instead minimized the empirical GE, in essentially the same way that Lasso implements the empirical generalization error on the test set. However, since the empirical GE and the actual GE are different, especially in finite samples, the accuracy, convergence rate and divergence between the empirical and actual GE are of interest. By adapting and extending the VC inequality and the principle of SRM, we propose the following theorem that states the connection between the empirical GE minimizer and the structural risk minimizer for both the finite sample and asymptotic cases.
%
%
\begin{thm}
If $\sup|\mathcal{R}_{n_s}(b)-\mathcal{R}(b)|\overset{\mathbf{P}}{\rightarrow}0$ for the extremum estimator $b$, the following Bahr-Esseen bound for the empirical GE holds with probability at least $\varpi(1-1/n_t)$, $\forall\varpi\in\left(0,1\right)$.
\begin{equation}
\mathcal{R}_{n_s}(b|X_{s},Y_{s}) \leqslant \overline{M} + \varsigma,
\label{thm1}
\end{equation}
where $\mathcal{R}_{n_s}(b|X_{s},Y_{s})$ is the empirical risk of $b$ on the test set,
\[
\overline{M}=\frac{\mathcal{R}_{n_t}(b|X_{t},Y_{t})}{(1-\sqrt{\epsilon})},
\]
\[
\varsigma = \frac{\sqrt[p]{2} \cdot \tau \left(\mathbb{E}\left[\mathrm{Loss}(y_{i},\widehat{m}(x_{i},b_{train}))\right]\right) } {\sqrt[p]{1-\varpi}\cdot n_s^{1-1/p}},
\]
where $p$ is a number strictly larger than $1$ and $\tau$ has been defined in Lemma~2.
\end{thm}
Thus, we immediately have the following corollary.
%
%
\begin{cor}
Based on Theorem 1, as $\widetilde{n}\rightarrow\infty$ the empirical GE minimizer and the structural risk minimizer converge to the same limit.
\end{cor}
Theorem~1 and Corollary~1 establish a foundation to study the control of model complexity, including the use of Lasso as an empirical GE minimizer, and also prove that, from a distribution-free and model-free perspective, SRM is asymptotically equivalent to empirical GE minimization. By using the bound in (\ref{thm1}), it is possible to quantify the difference between the effects of SRM and empirical GE minimization, and it is also possible to derive a confidence bound for the difference between SRM and empirical GE minimization.

SRM and empirical GE minimization offer a new angle to control model complexity and model selection, especially for Lasso. As shown in the CV-Lasso algorithm in Appendix~2, Lasso returns a vector of $b_{\lambda}$ for each $\lambda$. Larger values for $\lambda$ are mapped to a smaller VC dimension $h$ or $p$, referred to as the `admissible structure' of the model \citep{vc71b}. Among the list of models returned by Lasso, each different $p$ (VC dimension) parameterizes a generalization error. By picking the model with minimal empirical GE from $\left\{ b_{\lambda}\right\}$, both SRM and empirical GE minimization guarantee that the model chosen by Lasso has the best generalization ability.

\section{Generalization ability and consistency of Lasso-type model selection}

Section~2 shows that empirical GE minimization reduces overfitting, implying the estimator has a lower generalization error on out-of-sample data. In this section, we implement empirical GE minimization on linear regression with an $\mathcal{L}_{1}$ penalty. We show that, compared to the corresponding extremum estimator without penalty the $\mathcal{L}_{1}$-penalized extremum estimator, such as Lasso, minimizes the GE, improves out-of-sample performance, and controls for the overfitting problem. Moreover, the trade-off between in-sample and out-of-sample performance does not influence consistency. We also discuss the connection between the finite sample and asymptotic properties of the penalized extremum estimator.

The traditional route to prove consistency is through analyzing the properties of the extremum estimator in the training set as $n\rightarrow\infty$. However, to control overfitting and balance in-sample and out-of-sample fit, we need to consider the properties of estimators on both the training and the test sets. Thus, we derive the finite sample and asymptotic properties following the scheme outlined in Figure~2.

\begin{figure}
\centering
\includegraphics[width=0.6\paperwidth]{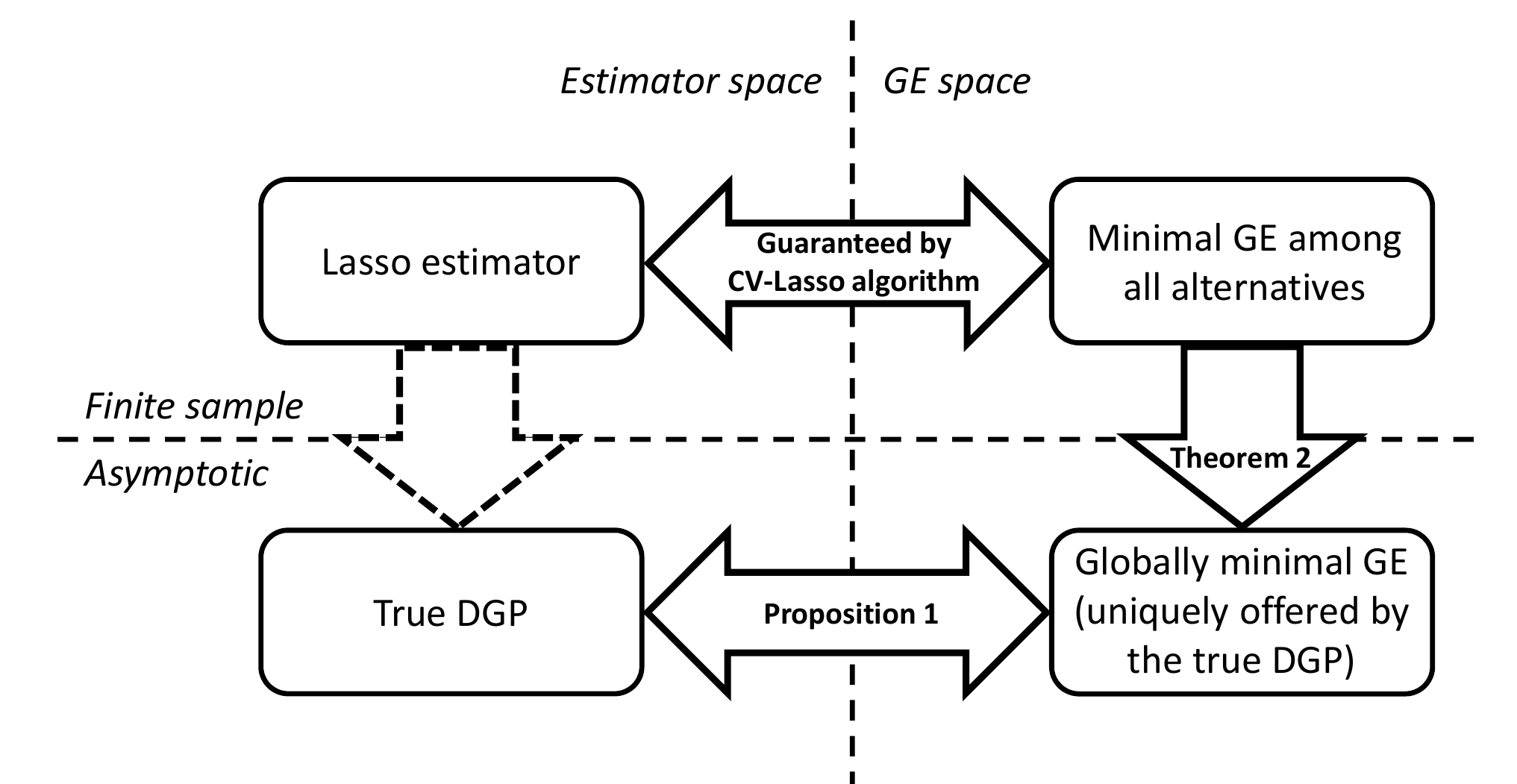}

\caption{Outline of proof strategy}
\end{figure}

An outline of our proof strategy is shown in Figure~2. Instead of working in the space of estimators, we reformulate the consistency problem in the space of generalization error. We show that empirical GE minimization not only controls overfitting and improves finite-sample performance, but also it helps us to find the true DGP asymptotically. We denote $b_{Lasso}$ as the model with the minimal GE among the alternatives. Lasso bijectively maps $b_{Lasso}$ to the minimal GE on the test set, defined as
$\tau:\,b_{Lasso}\rightarrow\min_{b_{\lambda}}\{\mbox{GEs of potential models}\}$.
To ensure GE minimization guides us towards the true DGP, we need first to prove that the mapping $\tau$ also bijectively assigns $\beta$ to the minimal GE in population, and second that if
\[
\min_{b\in b_{\lambda}}
\frac{1}{n_s}\sum_{i=1}^{n_s}\Vert Y_{s}-X_{s}b\Vert_{2}^{2}
\rightarrow
\min_b \int\Vert y-x^{T}b\Vert_{2}^{2}~\mathrm{d}F(x,y),
\]
then
\[
b_{Lasso}
\Leftrightarrow
\min\{\mbox{GEs of potential models}\}
\overset{\mathbf{P}}{\rightarrow}
\min_b \int\Vert y_{s}-x_{s}^{T}b\Vert_{2}^{2}~\mathrm{d}F(x,y)
\Leftrightarrow
\beta
\]
or, in other words, that $b_{Lasso}$ is consistent. This approach applies not only to the Lasso but also to other estimators designed to control overfitting or implement model selection.

\subsubsection*{Assumptions and identification }

At the outset, we stress that each variable in $(X,Y)$ must be standardized before implementing the Lasso. Without standardization, the Lasso algorithm may be influenced by the magnitude (units) of the variables.\footnote{An intuitive explanation \citep{tibshirani96} is that Lasso shrinks the absolute value of each $b_{i}$ by the same $\vert\lambda\vert$. Without standardization, variables with a smaller scale will have larger coefficients and are less likely to be dropped than variables with a larger scale and smaller coefficients.} After standardization, of course, $X$ and $Y$ unit- and scale-free.

To ensure the $\mathcal{L}_{2}$ consistency of Lasso, we require the following four assumptions.
\begin{description}
\item [{A1}] The true DGP is $Y=X\beta+u$.
\item [{A2}] $\mathbb{E} \left( u^T X \right) = \mathbf{0}$.
\item [{A3}] The true DGP is unique: no variable with a non-zero $\beta_{i}$ can be represented by a linear combination of any other variable in $X$.
\item [{A4}] Both the training set and the test set are i.i.d.\ from the same population.
\end{description}
The assumptions warrant a few comments. A1 restricts attention to linear regression models. A2 is the usual exogeneity condition. A3 is necessary for model selection; otherwise there may exist another model that is not statistically different from the population DGP. Note that A3 allows for linear dependence for the regressors with zero coefficients, but it does not allow any linear dependence to affect the true DGP. Thus, A3 is weaker than the typical assumption made for OLS that rules out perfect collinearity for all regressors. Lastly, A4 implies that we focus on the i.i.d.\ case in this paper. If A4 is not satisfied, a sample could consist of data from two completely different DGPs and Lasso generally cannot select a single model to represent two different DGPs.\footnote{In another paper we propose a `clustered Lasso' algorithm to deal with the non-i.i.d. case.}

Under assumptions A1 to A4, we show that the true DGP is the most generalizable model, yielding Proposition~1.
%
%
\begin{prop}
Under assumptions A1 to A4, the true DGP, $Y = X\beta + u$, is the one and only one offering the minimal generalization error as $\widetilde{n}\rightarrow\infty$.
\end{prop}
Proposition~1 states that there is a bijective mapping between $\beta$ and the globally minimal GE in the population. If A2 or A3 are violated, there may exist variables in the sample that render the true DGP not to be the most generalizable model. The Lasso algorithm picks the model with the minimal GE. As a result, we also need to prove that, when the sample size is `large' enough, the true DGP is included in the list of models from which Lasso selects. This is shown in Proposition~2.
%
%
\begin{prop}
Under assumptions A1 to A4 and Proposition~1, there exists at least one $\widetilde{\lambda}$ such that $\lim_{\widetilde{n}\rightarrow\infty}b_{\widetilde{\lambda}}=\beta$.
\end{prop}
In Lemmas~1 and~2 and in Theorem 1, we show that minimizing the empirical GE guarantees that the minimal empirical GE in the sample converges to the minimum GE in the population as $\widetilde{n}\rightarrow\infty$. We also show in Propositions~1 and~2 that $\beta$ uniquely offers the minimum GE in the population and is feasible for some $\widetilde{\lambda}$. Hence, the minimal GE in the sample converges to the population minimum GE, which is offered by $\beta$ uniquely, at some $\widetilde{\lambda}$.

Note that Lasso-type estimation is equivalent to the constrained minimization of a loss function. In Figure~3, the diamond-shape feasible area is determined by the $\mathcal{L}_1$ penalty, $b_{Lasso}$ refers to the Lasso estimates, `beta' refers to $\beta$, and $b_{OLS}$ refers to the OLS estimates. Different values for $\lambda$ imply different boundaries for the feasible area of the constrained minimization; the feasible area gets smaller as value of $\lambda$ gets larger. Hence, one of three cases may occur: (1) for a small value of $\lambda$, $\beta$ remains in the feasible area (under-shrinkage); (2) for $\lambda=\lambda^{*}$, $\beta$ is located precisely on the boundary of the feasible area (perfect-shrinkage); (3) for a large value of $\lambda$, $\beta$ is outside of the feasible area (over-shrinkage). In cases (1) and (2), the constraints become inactive as $\widetilde{n}\rightarrow\infty$,
so $\lim_{\widetilde{n}\rightarrow\infty}b_{\lambda}=\lim_{\widetilde{n}\rightarrow\infty}b_{OLS}=\beta$.
However, in case (3), $\lim_{\widetilde{n}\rightarrow\infty}b_{\lambda}\neq\beta$. Therefore, $\lim_{\widetilde{n}\rightarrow\infty}b_{\widetilde{\lambda}}=\beta$,
$\forall\widetilde{\lambda}\in\{\lambda|0\leqslant\lambda\leqslant\lambda^{*}\}$.

\begin{figure}
\centering
\includegraphics[width=0.22\paperwidth]{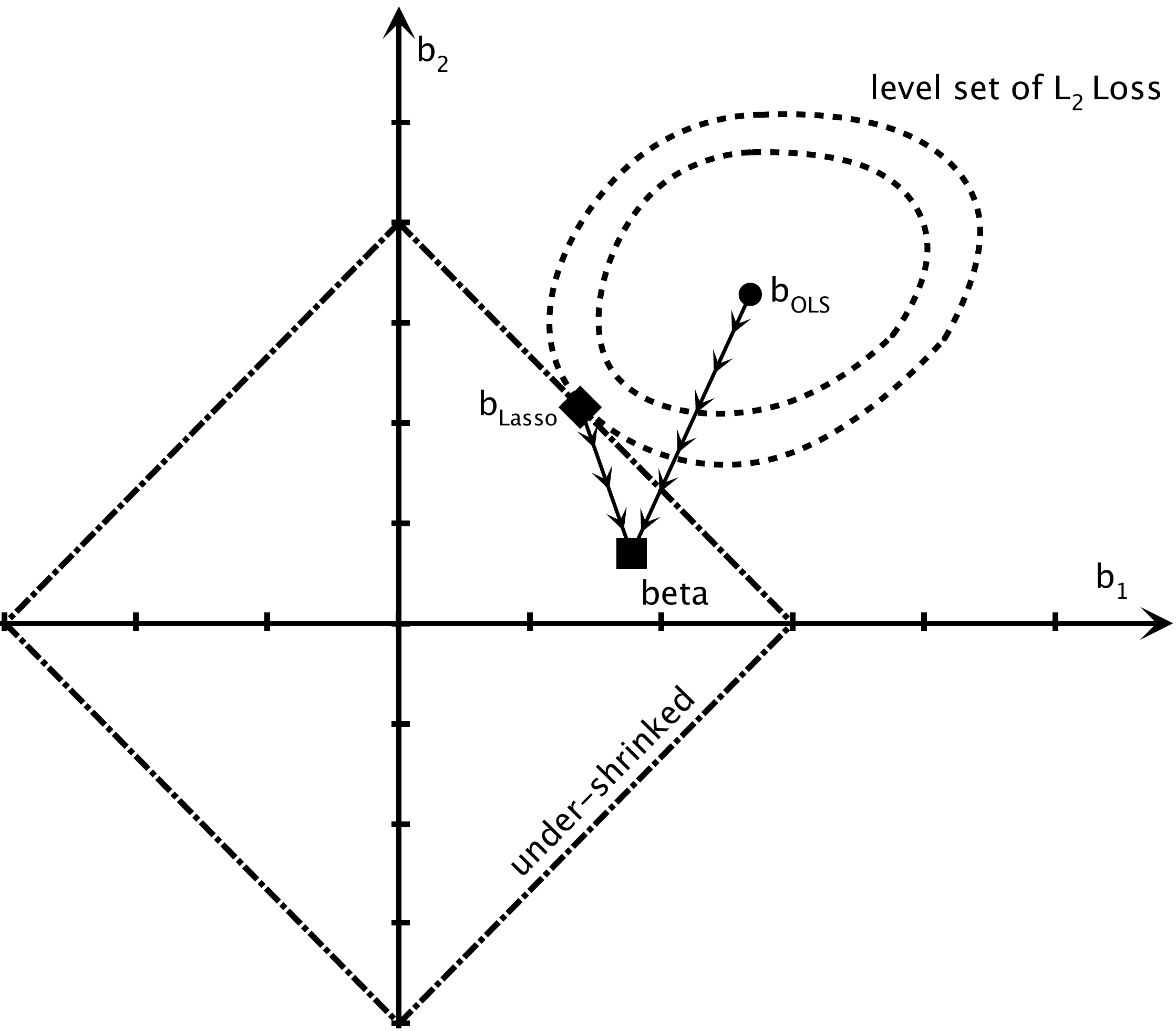}
\includegraphics[width=0.22\paperwidth]{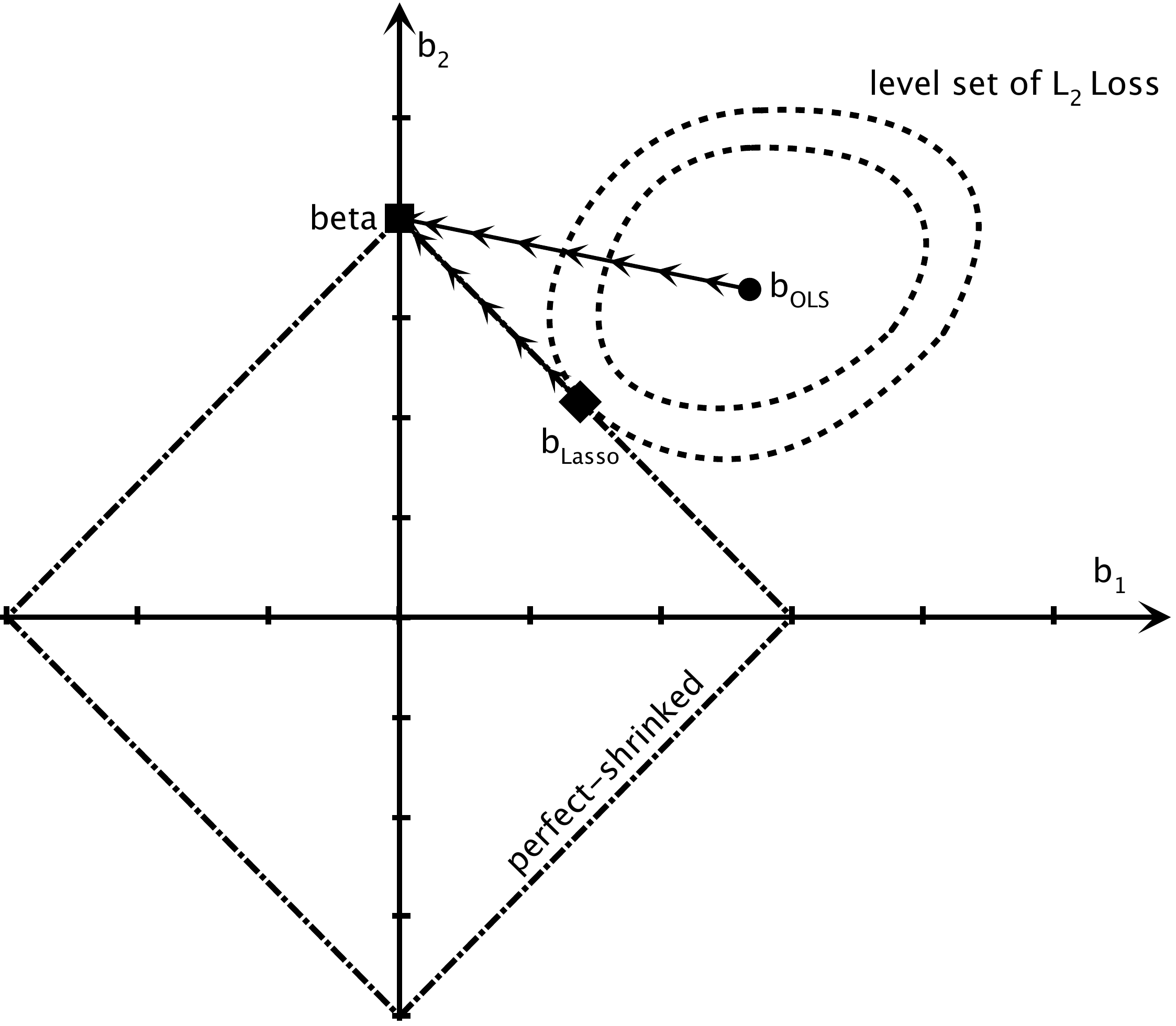}
\includegraphics[width=0.22\paperwidth]{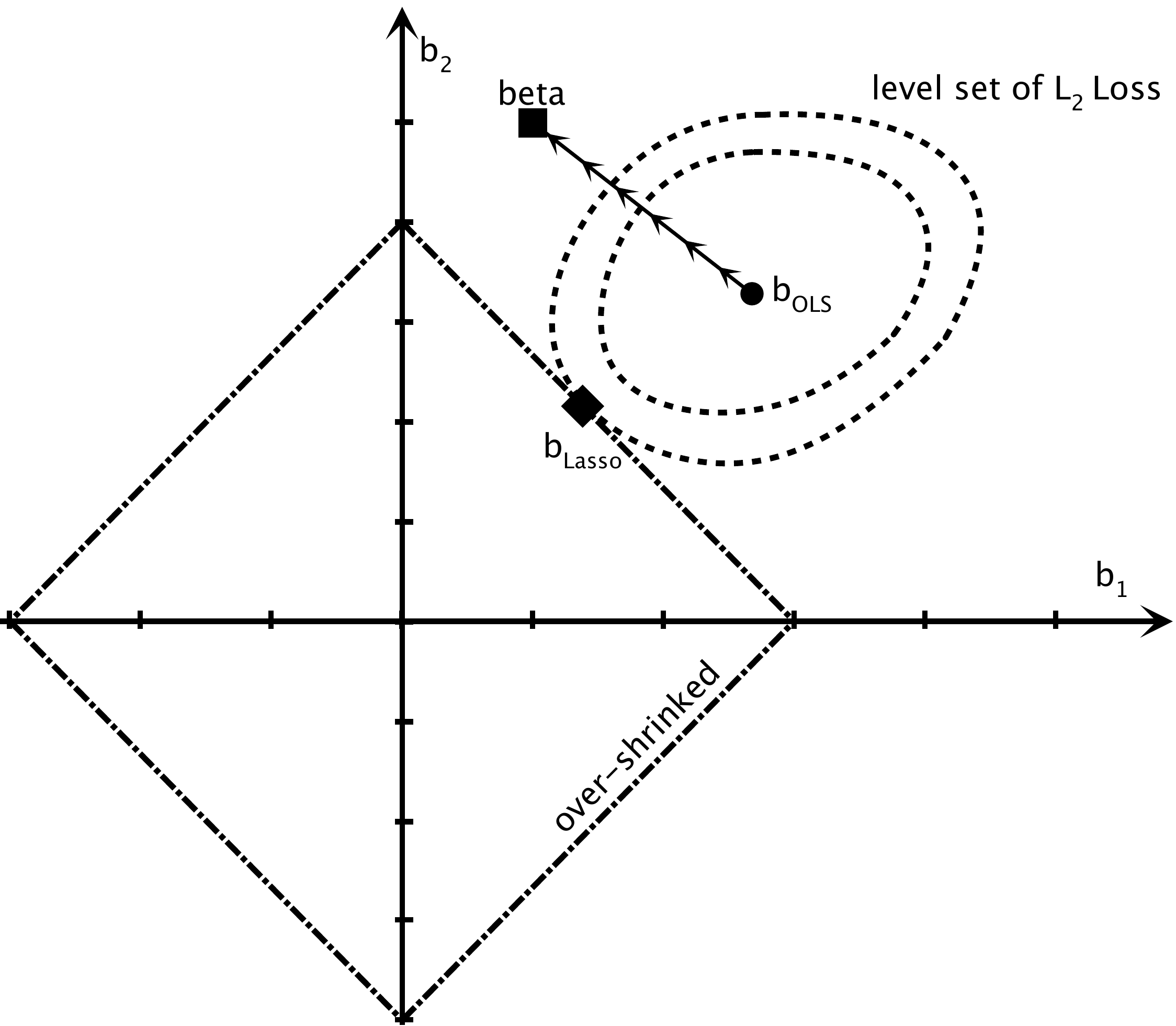}

\caption{Solutions of Lasso and $\beta$}
\end{figure}

As we show above, in practice we do not observe $\lambda^{*}$ a priori. The missing part of the puzzle is to find $\lambda\rightarrow\widetilde{\lambda}$ as $n\rightarrow\infty$. Thus, given Propositions~1, 2~and Theorem~1, we now show that empirical GE minimization guarantees the model selected by Lasso asymptotically converges in $\mathcal{L}_{2}$ to the true DGP, completing the `transformation' idea from Figure~2.
%
%
\begin{thm}
Based on Theorem~1, Propositions~1 and~2, under assumptions A1 to A4, the following bound holds with probability $\varpi(1-1/n_{t})$
\begin{equation}
\frac{1}{n_{s}}\Vert X_{s}b_{train}-Xb_{Lasso}\Vert _{2}^{2}
\leqslant
\left(\frac{1}{n_{t}}\frac{\Vert e_{t}\Vert_{2}^{2}}{1-\sqrt{\epsilon}}
-\frac{1}{n_{s}}\Vert e_{s}\Vert _{2}^{2}\right)
+\frac{4}{n_{s}}\Vert e_{s}^{T}X_{s}\Vert_{\infty}\Vert b_{train}\Vert _{1}+\varsigma
\label{thm2}
\end{equation}
where $b_{train}$ is the extremum estimator based on the training set and we define $e_{t}=Y_{t}-X_{t}b_{train}$ and $e_{s}=Y_{s}-X_{s}b_{train}$.
\end{thm}

Theorem~2 holds if $\lambda$ is tuned by validation. Moreover, the VC inequality and Theorem~2 can be generalized to the scenario where $\lambda$ is tuned by $K$-fold cross-validation. When Lasso is implemented by $K$-fold cross-validation, the sample is partitioned into $K$ equal-sized folds. If $K=2$, the theoretical result for $K$-fold cross-validation is identical to Theorem~2.\footnote{The $K=2$ case is also called holdout-validation.} For $K\geqslant 3$, we have $K$ different test sets for tuning $\lambda$ and $K$ different training set for estimation. Denote the $q$\textsuperscript{th} training set as $\left(X_t^q,Y_t^q\right)$, the $q$\textsuperscript{th} test set as $\left(X_s^q,Y_s^q\right)$, the extremum estimator estimated from the $k$\textsuperscript{th} training set as $b_{train}^k$, the sample size for each test set as $n_s$ and the sample size for each training set as $n_t$.

Denote $\mathrm{argmax}_{k,q}{\mathcal{R}_{n_s}(b_{train}^k|X_s^q,Y_s^q)}$ as $k^*$ and $q^*$. To simplify notation, we denote the extremum estimator for the worst case, $b_{train}^{k^*}$, by $\thickbar{b}_{train}$, $\varsigma_{k^*}$ by $\thickbar{\varsigma}$, $\epsilon_{k^*}$ by $\thickbar{\epsilon}$, and $\varpi_{k^*}$ by $\thickbar{\varpi}$. Hence, for any $k\;\mbox{and}\;q\in\left[1,K\right]$,
\begin{align*}
\mathcal{R}_{n_s}(b_{train}^k|X_s^q,Y_s^q) & \leqslant
\mathcal{R}_{n_s}\left(\thickbar{b}_{train}|X_s^{q^*},Y_s^{q^*}\right) \\
&\leqslant \mathcal{R}_{n_t}\left(\thickbar{b}_{train}|X_t^{q^*},Y_t^{q^*}\right)\left(1-\sqrt{\thickbar{\epsilon}}\right)^{-1}
+\thickbar{\varsigma}
\end{align*}
In this equation, we define the `worst case' to be where the GE among $K$ validations, $\mathcal{R}_{n_s}\left(b_{train}^{k}|X_s^{q},Y_s^{q}\right)$, is the largest among all validations.

Here we propose the following probabilistic bound for the Lasso tuned by $K$-fold cross-validation.

%
%
\begin{cor}
Based on Theorem~1 and Propositions~1 and~2, under assumptions A1 to A4, the following bound holds for the $K$-fold cross-validated Lasso with probability $\thickbar{\varpi}(1-1/n_{t})$
\begin{align*}
\frac{1}{K}\sum_{q=1}^K \frac{1}{n_s}\left\Vert X_s^q\thickbar{b}_{train} -X_s^qb_{Lasso}\right\Vert _{2}^{2} {} &
\leqslant
\left\vert\frac{1}{n_t}\frac{\left\Vert\overline{e_t}\right\Vert_{2}^{2}}{1-\sqrt{\overline{\epsilon}}}
-\frac{1}{K}\sum_{q=1}^K\frac{1}{n_s}\left\Vert\overline{e_s^q}\right\Vert_{2}^{2}\right\vert \\
& +\frac{1}{K}\sum_{q=1}^K\frac{4}{n_s}\left\Vert\left(\overline{e_s^q}\right)^TX_s^q\right\Vert_{\infty}
\left\Vert\thickbar{b}_{train}\right\Vert_{1}
+\overline{\varsigma}.
\end{align*}
where $\overline{e_t}$ is the largest training error of $\thickbar{b}_{train}$ on the training set, and $\overline{e_s^q}$ is the GE of $\thickbar{b}_{train}$ on the $q$\textsuperscript{th} test set.
\end{cor}

Using Theorem~2, Theorem~3 proves that Lasso is consistent for the $n_{t}\geqslant p$ case.
%
%
\begin{thm}
Based on Theorem~2, under assumptions A1 to A4, for $n_t \geqslant p$, the following bound holds with probability $\varpi(1-1/n_{t})$
\begin{equation}
\Vert b_{train}-b_{Lasso}\Vert _{2} \leqslant
\sqrt{\left\vert\frac{1}{\rho n_{t}}\frac{\Vert e_{t}\Vert_{2}^{2}}{(1-\sqrt{\epsilon})}
               -\frac{1}{\rho n_{s}}\Vert e_{s}\Vert_{2}^{2}\right\vert}
+\sqrt{\frac{4}{\rho n_{s}}\Vert e_{s}^{T}X_{s}\Vert_{\infty}\Vert b_{train}\Vert_{1}}
+ \left(\frac{\varsigma}{\rho}\right)^{\frac{1}{2}} \label{thm3}
\end{equation}
where $\rho$ is the minimal eigenvalue of $X^{T}X$ and $b_{train}$ is the OLS estimator. As a result, based on this bound, both OLS and the Lasso estimator converge in the $\mathcal{L}_{2}$ norm asymptotically to the true DGP if \textup{$\lim_{n\rightarrow\infty}p/\widetilde{n}=0$}.
\end{thm}

For $n_t\geqslant p$, if Lasso is tuned by cross-validation, a slightly different probabilistic bound can be derived based on Theorem~3, Corollary 2 and Theorem 1, as follows.
%
%
\begin{cor}
Based on Theorem~3, Corollary~2 and Theorem 1, under assumptions A1 to A4, for $n_t \geqslant p$, the following bound holds with probability $\varpi(1-1/n_{t})$
\begin{align*}
\frac{1}{K}\sum_{q=1}^K \left\Vert \thickbar{b}_{train} - b_{Lasso}\right\Vert _{2}^{2} &
\leqslant
\left\vert\frac{1}{n_t\cdot\overline{\rho}}\frac{\left\Vert\overline{e_t}\right\Vert_{2}^{2}}{1-\sqrt{\overline{\epsilon}}}
-\frac{1}{K}\sum_{q=1}^K\frac{1}{n_s\cdot\overline{\rho}}\left\Vert \overline{e_s^q}\right\Vert_{2}^{2}\right\vert\\
& +\frac{1}{K}\sum_{q=1}^K\frac{4}{n_s\cdot\overline{\rho}}\left\Vert \left(\overline{e_s^q}\right)^TX_s^q\right\Vert_{\infty}\left\Vert\thickbar{b}_{train}\right\Vert_{1}
  +\frac{\overline{\varsigma}}{\overline{\rho}}
\end{align*}
where $\overline{\rho}$ is defined as $\min\left\{\rho_k\vert\rho_k\mbox{ is the minimal eigenvalue of} \left(X^k_s\right)^{T}X^k_s, \forall k\right\}$ and  $\thickbar{b}_{train}$ is  is the OLS estimator that caused the largest GE in $K$ validations. As a result, based on this bound, both OLS and the Lasso estimator converge in the $\mathcal{L}_{2}$ norm asymptotically to the true DGP if \textup{$\lim_{n\rightarrow\infty}p/\widetilde{n}=0$}.
\end{cor}

Since OLS requires that $X^T X$ is of full-rank, it cannot be directly implemented in cases where $p>n$. In such cases, the extremum estimator $b_{train}$ must satisfy $\mathrm{dim}\left(b_{train}\right)\leqslant n$.  Hence, the extremum estimator for $p>n$ may be implemented by forward selection regression (FSR) without constraining $\Vert b\Vert_1$. To avoid including too many variables, FSR is designed to stop when $\mathrm{corr}\left(u,x_i\right)$ is less than some preset number for all $x_i$ that are not chosen by forward selection. To be specific, as shown by \citet{efronall04}, Lasso may be seen as a forward selection regression with an $\mathcal{L}_1$ norm constraint.\footnote{The method of solving Lasso by forward selection is the least angle regression (LARS). For details of LARS and its consistency, see \citet{efronall04} and \citet{zhang10}.} \citet{zhang10} shows (algorithm~2), that FSR finds the combination of variables, $\mathcal{H}$, that minimizes the regression training error under the restriction that the number of variables in $\mathcal{H}$ is less or equal to $\min\left(n_t,p\right)$, which is similar to Lasso. Moreover, Zhang shows that FSR is a greedy algorithm that may result in overfitting in finite samples. He also shows that FSR is $\mathcal{L}_2$-consistent under the sparse eigenvalue condition \citep{bickeletal09,meinshausenyu09}. Therefore, in cases where $p>n$, we set the FSR estimator to be $b_{train}$. In Theorem~4, we show that the Lasso reduces the overfitting of FSR and is $\mathcal{L}_2$-consistent for the $p>n$ case by importing the sparse eigenvalue condition from \citet{bickeletal09,meinshausenyu09}---see the proof of Theorem~4 in Appendix~1 for the details.
%
%
\begin{thm}
Based on Theorem~1, Theorem~2 and Corollary~2, under assumptions A1 to A4 and the restricted eigenvalue assumption, for the case $p>n_{t}$, the following bound holds with probability $\varpi(1-1/n_{t})$
\begin{align}
\Vert b_{train}-b_{Lasso}\Vert_{2} & \leqslant
\sqrt{\left|\frac{1}{\rho_{re}n_{t}}\frac{\Vert e_{t}\Vert_{2}^{2}}{(1-\sqrt{\epsilon})}
          - \frac{1}{\rho_{re}n_{s}}\Vert e_{s}\Vert_{2}^{2}\right|} \nonumber \\
& + \sqrt{\frac{4}{\rho_{re}n_{s}}\Vert e_{s}^{T}X_{s}\Vert_{\infty}\Vert b_{train}\Vert_{1}}
+ \left(\frac{\varsigma}{\rho_{re}}\right)^{\frac{1}{2}}
\label{thm4}
\end{align}
where $\rho_{re}$ is the minimum of the restricted eigenvalues of $X^{T}X$ and $b_{train}$ is the extremum estimator. As a result, both the Lasso and FSR estimator converge in the $\mathcal{L}_{2}$ norm to the true DGP if \textup{$\lim_{n\rightarrow\infty}\ln p/\widetilde{n}=0$}.
\end{thm}

For $n_t\leqslant p$, if Lasso is tuned by cross-validation, a slightly different probabilistic bound can be derived based on Theorem~4, Corollary 2 and Theorem 1, as follows.
%
%
\begin{cor}
Based on Theorem~1, Theorem~4 and Corollary~3, under assumptions A1 to A4 and the restricted eigenvalue assumption, for $n_t \geqslant p$, the following bound holds with probability $\varpi(1-1/n_{t})$
\begin{align*}
\frac{1}{K}\sum_{q=1}^K \left\Vert \thickbar{b}_{train} - b_{Lasso}\right\Vert _{2}^{2} &
\leqslant
\left\vert\frac{1}{n_t\cdot\overline{\rho}}\frac{\left\Vert \overline{e_t}\right\Vert_{2}^{2}}{1-\sqrt{\overline{\epsilon}}}
-\frac{1}{K}\sum_{q=1}^K\frac{1}{n_s\cdot\overline{\rho}}\left\Vert \overline{e_s^q}\right\Vert_{2}^{2}\right\vert\\
& +\frac{1}{K}\sum_{q=1}^K\frac{4}{n_s\cdot\overline{\rho}}\left\Vert \left(\overline{e_s^q}\right)^TX_s^q\right\Vert_{\infty}\left\Vert\thickbar{b}_{train}\right\Vert_{1}
  +\frac{\overline{\varsigma}}{\overline{\rho}}
\end{align*}
where $\overline{\rho}$ is defined as $\min\left[\widetilde{\rho}_k\vert\widetilde{\rho}_k\mbox{ is the minimal restricted eigenvalue of }\left(X^k_s\right)^{T}X^k_s, \forall k\right]$ and  $\thickbar{b}_{train}$ is  is the FSR estimator that caused the largest GE in $K$ validations. As a result, both the Lasso and FSR estimator converge in the $\mathcal{L}_{2}$ norm to the true DGP if \textup{$\lim_{n\rightarrow\infty}\ln p/\widetilde{n}=0$}.
\end{cor}

\begin{figure}
\centering
\includegraphics[width=0.4\paperwidth]{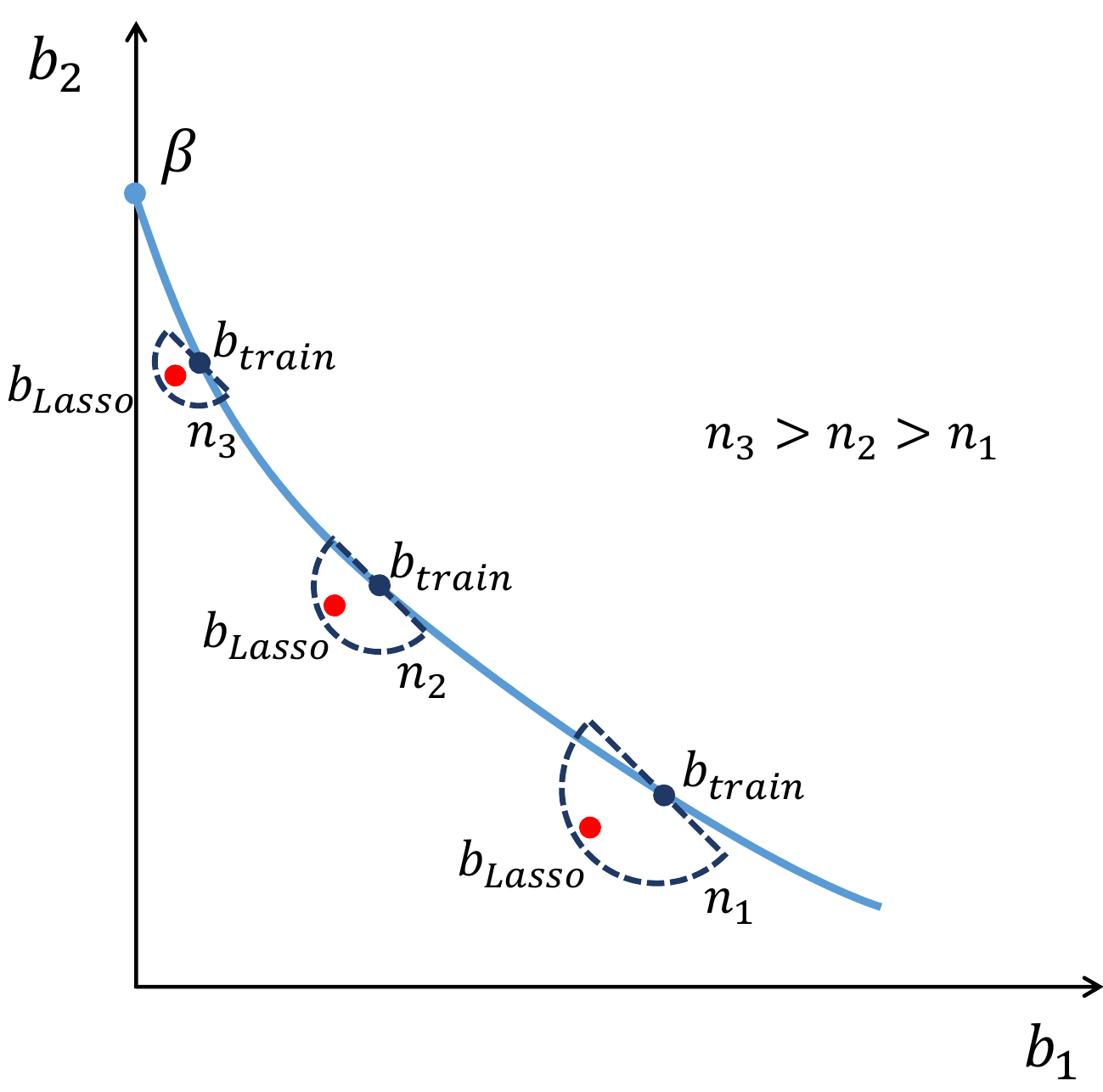}

\caption{Representation of $b_{train}$ and $b_{Lasso}$ convergence}
\end{figure}

Theorems~2 to~4 capture the relationship between the Lasso estimator $b_{Lasso}$ and the extremum estimator $b_{train}$, which is summarized in Figure~4. \citet{neweymcfadden94} show that the extremum estimator is consistent and converges to the true parameter $\beta$ as $n\rightarrow\infty$ under some regularity conditions. The line through the $b_{train}$'s to $\beta$ shows the corresponding path of convergence.\footnote{\citet{vc74b} also derive the necessary and sufficient condition for consistency of the extremum estimator, which they refer to as empirical risk minimization.}

Since Lasso is implemented to reduce the GE, $b_{train}$ and $b_{Lasso}$ will typically be numerically different. Theorems~2 to~4 show that, with probability $\varpi(1-1/n_{t})$, the $\mathcal{L}_{2}$ difference between $b_{train}$ and $b_{Lasso}$ is bounded by the sum of three terms: overfitting caused by the extremum estimator (the first RHS term (\ref{thm3}) and (\ref{thm4})\footnote{The first RHS term in each of these two equations is also related to the difference between the training error and the testing error.}), error due to $e_s^TX/n_s\neq 0$ in the test set (the second RHS term in (\ref{thm3}) and (\ref{thm4})) and sampling error on the test set (the last RHS term in (\ref{thm3}) and (\ref{thm4}))\footnote{The last RHS term in each of these two equations is derived from the Hoeffding inequality which is used in the proof of Theorem~1.}. Hence, as shown in Figure~4, the Lasso estimator (the empirical GE minimizer) generally does not lie on the convergence path of the extremum estimator. However, Theorems~2 to~4 show that the deviation of $b_{Lasso}$ from the convergence path is bounded. To be specific, $b_{Lasso}$ always lies within the feasible area parameterized by $\lambda\Vert b\Vert_1$. Graphically, $b_{Lasso}$ lies within an $\epsilon$-ball centered on $b_{train}$ with radius given by the RHS of (\ref{thm3}) or (\ref{thm4}). As shown in Figure~4, $b_{Lasso}$ always lies within the bounds of the $\epsilon$-ball feasible area shown by the dashed $45^\circ$-offset semi-circles. As $n/p$ increases, the $\epsilon$-ball becomes smaller, the Lasso estimator gets closer to the extremum estimator, and both converge to $\beta$.

By implementing the empirical GE minimizer, Lasso reduces overfitting and increases generalization ability. Hence, we show the connection between minimizing GE and asymptotic performance. This justifies using Lasso for model selection: it is consistent if all the assumptions are satisfied and even if an assumption is not satisfied in practice, it still offers a model with maximal generalization ability. The maximal generalization ability is typically considered useful for empirical research, such as policy analysis, since it makes the performance of the estimated model stable when applied to out-of-sample data.

\subsubsection*{Connection to previous work}

Our approach establishes $\mathcal{L}_{2}$ consistency for Lasso from a different perspective, as well as verifying, generalizing or complementing the results of following papers.

We extend and broaden the scope of VC theory and SRM. \citet{vc71b} originally propose the SRM principle in the context of group classification models. By balancing the in-sample and out-of-sample fit, SRM finds the best `off-shore' classification algorithm. Alongside our transform strategy and Theorem~2, SRM can be applied to study the properties of numeric algorithms and estimators, general proofs of consistency proof functional spaces, and so on. In another paper we reveal the full power of SRM by extending the results to general spaces of functionals.

Our approach offers a new angle from which to view OLS and linear function approximation. As we show in introducing VC theory, the training error will be close to the generalization error if $n/p$ is very large. Hence, OLS may be viewed as a special case of SRM, in which training error is considered approximately identical to generalization error. Moreover, typically we may approximate any DGP with linear regression because any analytic function can approximated by an infinite series of polynomials, at least locally. However, in practice this idea encounters three problems: (1) it is impossible to formulate infinite series in empirical research, (2) for high-dimensional data we need to decide which variable to include, and (3) it collapses immediately if the DGP is non-analytic. Thus, SRM and empirical GE minimization implemented by Lasso, offer a new angle on approximation: we approximate the generalization ability of true DGP. If in population the GE of the true DGP can be distinguished from other models, minimizing the GE will guide estimation to the true DGP eventually. Even if the DGP is not well-defined, asymptotically minimizing the GE will provide an approximation that improve model performance on out-of-sample data.

\citet{zhaoyu06}, \citet{meinshausenyu09}, and \citet{knightfu00} derive a necessary condition (and a relaxed version) for probabilistic consistency of Lasso, called the \textit{irrepresentable condition}, by defining $X=[X_{1},X_{2}]$, where the $X_{1}$ are elements in the true DGP and the $X_{2}$ are redundant. The condition claims that \textit{Lasso is consistent in probability only if} $\Vert(X_{1}^{T}X_{1})^{-1}X_{1}^{T}X_{2}\,\mathrm{sign}(b)\Vert_{1}<1$. Intuitively, this condition implies that if we regress redundant variables on any variable in the true DGP, the norm of coefficient parameter cannot be larger than~$1$ as $\Vert(X_{1}^{T}X_{1})^{-1}X_{1}^{T}X_{2j}\,
\mathrm{sign}(b)\Vert_{1}=\sum_{i=1}^p|\mathrm{corr}(X_{1i},X_{2j})|<1$. Our assumptions are less restrictive since A3 only requires that the true DGP is unique.

\citet{shao97} compares the performance of model selection across AIC, BIC, cross-validation and other methods, and proposes conditions to make generalized information criterion (GIC) and cross-validation consistent in model selection. $K$-fold cross-validation is consistent if the set of alternative models contains at least one correct model with a fixed dimension. By introducing VC theory, our work compliments and extends Shao's condition in two ways. Firstly, we introduce the finite sample property of a method to implement SRM. Second, GIC and Lasso share a similar condition for consistency in terms of penalizing an over-complicated model. Our condition is consistent with Shao's since we implement Lasso by cross-validation.

Lastly, some researchers have modified the Lasso to deal with specific scenarios, such as adaptive Lasso\citep{zou06}, relaxed Lasso\citep{meinshausen07}, and group Lasso \citep{friedman10}. It is straightforward to extend our framework and results to these algorithms.

\section{Simulation Study}

We illustrate our theoretical results using simulations. We assume the outcome variable $y$ is generated by the following DGP:
\[
y = X'\beta+u = X_{1}'\beta_{1} + X_{2}'\beta_{2} + u
\]
where $X=\left(x_{1},\cdots,x_{p}\right)\in\mathbb{R}^{p}$ is generated by a multivariate Gaussian distribution with zero mean, $var(x_{i}) = 1$, $corr(x_{i},x_{j}) = 0.9,\forall i,\thinspace j$, $\beta_{1}=\left(2,4,6,8,10,12\right)^T$ and $\beta_{2}$ is a $(p - 6)$-dimensional zero vector. $u$ is generated from a Gaussian distribution with zero mean and unit variance. Here $x_{i}$ doesn't cause $x_{j}$ and no causal relation exists between $u$ and $x_{i}$.

We set the sample size at 250 and $p$ at four values: 200, 250, 300, 500. In each case, we repeat simulation 50 times. In each simulation, we apply the Lasso algorithm to find the estimate of $\beta$ and calculate its distance to the true value, the generalization error, and the in-sample/out-of-sample goodness-of-fit measure $GR^{2}$. As a comparison, we also apply OLS for the $n\geqslant p$ cases or the forward selection regression (FSR) algorithm for the $n<p$ cases.

Boxplots (see Appendix~3) show the estimates of all coefficients in $\beta_{1}$ (labeled $b_1$ to $b_6$) along with the four worst estimates of coefficients in $\beta_{2}$ (labeled $b_7$ to $b_{10}$), where `worst' refers to the estimates with the largest bias. The Lasso and OLS/FSR estimates and histograms of the $GR^{2}$ are reported for each case, respectively, in Figures 5--8 (Appendix~3). Finally, the distance between the estimates and the true values, the generalization error, and $GR^{2}$ (averages across the 50 simulations) are reported in Table~1 for all four cases for $p$.

When $n>p$, as we can see from the boxplot in Figure~5, both Lasso and OLS perform well. All the coefficient estimates are centered around the corresponding true values, and the deviations are relatively small. However, Lasso outperforms OLS for the estimates of $\beta_{2}$ in terms of having much smaller deviations. Indeed, a joint significance test ($F$ test) fails to reject the null hypothesis that all coefficients in $\beta_{2}$ are zero for the OLS estimates. As shown in Figure~5, the Lasso $GR^{2}$ is marginally larger than the OLS $GR^2$, but the differences are inconsequential.

When $n=p$, as shown in Figure~6, Lasso still performs well while it is apparent that OLS is biased and its deviations much larger. Also as shown in Figure~6, the Lasso $GR^{2}$ is clustered around 1 while the $GR^{2}$ for OLS takes on a range of values from 1 down to 0.2. This is evidence that OLS suffers from an overfitting problem.

When $n<p$, the regression model is not identified, OLS is infeasible, and we apply FSR. As shown in Figures~7 and~8, Lasso still performs well and correctly selects the variables with non-zero coefficients. In contrast, although FSR also correctly identifies the non-zero coefficients, its biases and deviations are much larger than for the Lasso. For the $p=500$ case shown in Figure~8 it is clear that the FSR estimates are unreliable. Generally speaking, overfitting is controlled well by Lasso (all the $GR^2$ are close to 1) whereas the performance of FSR is mixed, as reflected by the deteriorating $GR^2$ as $p$ increases. This suggests that, by imposing an $\mathcal{L}_1$ penalty on estimates, Lasso mitigates the overfitting problem and that the advantage of Lasso is likely to be more pronounced as $p$ increases.

\begin{table}[ht]
\caption{Average bias, training error, generalization error, in-sample $R^2$, out-of-sample $R^2$, and $GR^{2}$ for Lasso and OLS/FSR}
\centering

\begin{tabular}{l.....}
\toprule
Measure & \multicolumn{1}{r}{$p=200$}
        & \multicolumn{1}{r}{$p=250$}
        & \multicolumn{1}{r}{$p=300$}
        & \multicolumn{1}{r}{$p=500$} \\
\midrule
Bias\\
\hspace*{4mm}$b_{Lasso}$   & 0.7124 &   0.7382 &   0.7813 &   0.8713 \\
\hspace*{4mm}$b_{OLS/FSR}$ & 0.9924 &   9.7946 &   6.4417 &   6.3143 \\
Training error\\
\hspace*{4mm}Lasso         & 0.9007 &   0.8915 &   0.9048 &   0.8550 \\
\hspace*{4mm}OLS/FSR       & 0.2048 &   2.5856 & 374.9750 & 343.8078 \\
Generalization error\\
\hspace*{4mm}Lasso         & 1.1068 &   1.0998 &   1.1095 &   1.1396 \\
\hspace*{4mm}OLS/FSR       & 5.2109 & 525.4980 & 406.4791 & 359.5249 \\
 $R^2$, in-sample\\
\hspace*{4mm}Lasso         & 0.9994 &   0.9994 &   0.9994 &   0.9995 \\
\hspace*{4mm}OLS/FSR       & 0.9999 &   0.9985 &   0.7603 &   0.7821 \\
 $R^2$, out-of-sample\\
\hspace*{4mm}Lasso         & 0.9993 &   0.9993 &   0.9993 &   0.9993 \\
\hspace*{4mm}OLS/FSR       & 0.9968 &   0.6696 &   0.7534 &   0.7820 \\
$GR^{2}$\\
\hspace*{4mm}Lasso         & 0.9988 &   0.9988 &   0.9988 &   0.9987 \\
\hspace*{4mm}OLS/FSR       & 0.9967 &   0.6686 &   0.5728 &   0.6116 \\
\bottomrule
\end{tabular}
\end{table}

Table~1 reinforces the impressions from the boxplots and histograms. When $p=200$ OLS of course performs extremely well in terms of training error and more poorly in terms of generalization error while its $GR^2$ is very close to the Lasso value. For $p=250$ the performance of OLS deteriorates markedly in terms of bias, both the errors, and and out-of-sample fit, generating a corresponding fall in $GR^2$. For $n<p$ what is noteworthy is the stable performance of the Lasso relative to that of FSR. The training errors, generalization errors, and $GR^2$ are particulary poor for FSR, again illustrating the advantage of the Lasso in avoiding overfitting.

\section{Conclusion}

In this paper, by using SRM, we show that the maximization of generalization ability and model selection share the same algebraic and topological structure. If we address one, the other is also solved as well. This highlights the importance of generalization error minimization in model selection and parameter estimation. We establish the $\mathcal{L}_2$ consistency of Lasso-type model selection under assumptions (A1--A4) similar to those typically imposed on OLS. In this way, we ensure the Lasso is applicable to economic data, especially when big data is increasingly available. We propose the CV-Lasso algorithm which uses cross-validation to choose the $\mathcal{L}_1$ penalty parameter. The algorithm significantly reduces computation load and, thus, makes model selection in big data sets feasible. We also propose the generalized $R^2$, $GR^2$, to measure both in-sample and out-of-sample fitting.

We illustrate model selection consistency by simulations and demonstrate that the CV-Lasso algorithm has the potential to recover true DGPs if assumptions A1 to A4 are satisfied. It is clear that, under a range of settings, minimizing the generalization error picks the true DGP efficiently. In particular, the CV-Lasso algorithm strikes a good balance between in-sample and out-of-sample fitting, as indicated by $GR^2$. In another paper, we develop a new algorithm that is able to recover DGPs with a sophisticated hierarchical structure, which should find many potential applications in economics.

A potential concern is the reliability of the CV-Lasso algorithm when some of the assumptions A1--A4 do not hold. If one or more of the assumptions fail, consistency is not achievable. However, since the CV-Lasso algorithm is based on minimizing the generalization error, the model selected by the CV-Lasso algorithm will still offer good generalization ability. This is similar in spirit to the case of quasi-maximum likelihood, where the estimates may not be consistent but are still useful for inference.

There are two tuning parameters in implementing Lasso, $\lambda$ (the penalty parameter) and $K$ (the number of folds used in cross-validation). In this paper, we show that cross-validation selects a $\lambda$ that leads to consistent model selection and parameter estimation. Alternatively, the BIC may be used for the choice of $\lambda$. We conjecture that cross-validation is asymptotically equivalent to BIC in selecting $\lambda$. Simulations (not reported here) indicate that both cross-validation and BIC work well for selecting $\lambda$ in medium to large samples. In practice, the number of folds ($K$) in cross-validation is conventionally set at 5, 10, 20 or $n$ (leave one out). The choice of $K$ is of theoretical interest because it is related to the question of how much information is necessary for estimation and how much for validation. In another paper we provide some theoretical results surrounding the choice of $K$.

Our work sheds light not only on Lasso-type regressions, but also more generally on the applicability of model selection based on structural risk minimization, offering additional insight into the bias-variance trade-off. In this paper, we focus mainly on implementing Lasso-type regression through the minimization of generalization error. But Lasso could be implemented for maximum likelihood, functional regression, principle component analysis, decision trees and other estimation methods. Furthermore, the results here on Lasso-type model selection may be used together with other empirical methods. For instance, high dimensionality makes clustering hard because having lots of dimensions means that everything is `far away' from each other. High dimensionality is also an issue when estimation involves rejection sampling since the acceptance probability will keep shrinking with dimension and it becomes increasingly harder to find an appropriate enveloping distribution. In these cases, we may apply the CV-Lasso to pre-select variables for the following procedures.

\newpage{}
\section*{References}

\bibliographystyle{elsarticle-harv}
\bibliography{LASSOrefs}

\begin{thebibliography}{47}
\expandafter\ifx\csname natexlab\endcsname\relax\def\natexlab#1{#1}\fi
\expandafter\ifx\csname url\endcsname\relax
  \def\url#1{\texttt{#1}}\fi
\expandafter\ifx\csname urlprefix\endcsname\relax\def\urlprefix{URL }\fi

\bibitem[{Akaike(1973)}]{akaike73}
Akaike, H., 1973. Information theory and an extension of the maximum likelihood
  principle. In: Petrov, B.~N., Csaki, F. (Eds.), 2\textsuperscript{nd}
  International Symposium on Information Theory, Tsahkadsor, Armenia, USSR.
  Budapest: Akademiai Kaido, pp. 267--281.

\bibitem[{Bai and Ng(2008)}]{baing08}
Bai, J., Ng, S., 2008. Forecasting economic time series using targeted
  predictors. Journal of Econometrics 146~(2), 304--317.

\bibitem[{Bellman(1957)}]{bellman57}
Bellman, R.~E., 1957. Dynamic Programming. Rand {C}orporation research study.
  Princeton University Press.

\bibitem[{Belloni et~al.(2012)Belloni, Chen, Chernozhukov, and
  Hansen}]{belloniall12}
Belloni, A., Chen, D., Chernozhukov, V., Hansen, C.~B., 2012. Sparse models and
  methods for optimal instruments with an application to eminent domain.
  Econometrica 80~(6), 2369--2429.

\bibitem[{Belloni and Chernozhukov(2011)}]{belloni2011high}
Belloni, A., Chernozhukov, V., 2011. High dimensional sparse econometric
  models: An introduction. Springer.

\bibitem[{Bickel et~al.(2009)Bickel, Ritov, and Tsybakov}]{bickeletal09}
Bickel, P.~J., Ritov, Y., Tsybakov, A.~B., 2009. Simultaneous analysis of lasso
  and dantzig selector. The Annals of Statistics 37, 1705--1732.

\bibitem[{Breiman(1995)}]{breiman95}
Breiman, L., 1995. Better subset regression using the nonnegative garrote.
  Technometrics 37~(4), 373--384.

\bibitem[{Candes and Tao(2007)}]{candestao07}
Candes, E.~J., Tao, T., 2007. The dantzig selector: statistical estimation when
  $p$ is much larger than $n$. The Annals of Statistics, 2313--2351.

\bibitem[{Caner(2009)}]{caner09}
Caner, M., 2009. {Lasso}-type gmm estimator. Econometric Theory 25~(1),
  270--290.

\bibitem[{Chatterjee et~al.(2015)Chatterjee, Gupta, and
  Lahiri}]{chatterjeeall15}
Chatterjee, A., Gupta, S., Lahiri, S., 2015. On the residual empirical process
  based on the {ALASSO} in high dimensions and its functional oracle property.
  Journal of Econometrics 186~(2), 317--324.

\bibitem[{Cheng and Liao(2015)}]{chengliao15}
Cheng, X., Liao, Z., 2015. Select the valid and relevant moments: An
  information-based {Lasso} for gmm with many moments. Journal of Econometrics
  186~(2), 443--464.

\bibitem[{Chickering et~al.(2004)Chickering, Heckerman, and
  Meek}]{chickering04}
Chickering, D.~M., Heckerman, D., Meek, C., 2004. Large-sample learning of
  {B}ayesian networks is {NP}-hard. Journal of Machine Learning Research 5,
  1287--1330.

\bibitem[{De~Mol et~al.(2008)De~Mol, Giannone, and Reichlin}]{demolall08}
De~Mol, C., Giannone, D., Reichlin, L., 2008. Forecasting using a large number
  of predictors: Is bayesian shrinkage a valid alternative to principal
  components? Journal of Econometrics 146~(2), 318--328.

\bibitem[{Efron et~al.(2004)Efron, Hastie, Johnstone, and
  Tibshirani}]{efronall04}
Efron, B., Hastie, T., Johnstone, I., Tibshirani, R., 2004. Least angle
  regression. The Annals of statistics 32~(2), 407--499.

\bibitem[{Frank and Friedman(1993)}]{frankfriedman93}
Frank, I.~E., Friedman, J.~H., 1993. A statistical view of some chemometrics
  regression tools. Technometrics 35~(2), 109--135.

\bibitem[{Friedman et~al.(2010)Friedman, Hastie, and Tibshirani}]{friedman10}
Friedman, J., Hastie, T., Tibshirani, R., 2010. A note on the group lasso and a
  sparse group lasso. arXiv preprint arXiv:1001.0736.

\bibitem[{Friedman et~al.(1997)Friedman, Geiger, and Goldszmidt}]{friedman97}
Friedman, N., Geiger, D., Goldszmidt, M., 1997. Bayesian network classifiers.
  Machine Learning 29~(2-3), 131--163.

\bibitem[{Friedman et~al.(2000)Friedman, Linial, Nachman, and
  Pe'er}]{friedman00}
Friedman, N., Linial, M., Nachman, I., Pe'er, D., 2000. Using {B}ayesian
  networks to analyze expression data. In: Proceedings of the Fourth Annual
  International Conference on Computational Molecular Biology. RECOMB '00. ACM,
  New York, NY, USA, pp. 127--135.

\bibitem[{Fu(1998)}]{fu98}
Fu, W.~J., 1998. Penalized regressions: the bridge versus the {Lasso}. Journal
  of computational and graphical statistics 7~(3), 397--416.

\bibitem[{Heckerman et~al.(1995)Heckerman, Geiger, and
  Chickering}]{heckerman95}
Heckerman, D., Geiger, D., Chickering, D.~M., 1995. Learning {B}ayesian
  networks: The combination of knowledge and statistical data. Machine learning
  20~(3), 197--243.

\bibitem[{James and Stein(1961)}]{james1961}
James, W., Stein, C., 1961. Estimation with quadratic loss. In: Proceedings of
  the 4th Berkeley symposium on mathematical statistics and probability.
  Vol.~1. pp. 361--379.

\bibitem[{Kim and Swanson(2014)}]{kimswanson14}
Kim, H.~H., Swanson, N.~R., 2014. Forecasting financial and macroeconomic
  variables using data reduction methods: New empirical evidence. Journal of
  Econometrics 178, 352--367.

\bibitem[{Knight and Fu(2000)}]{knightfu00}
Knight, K., Fu, W., 2000. Asymptotics for {Lasso}-type estimators. Annals of
  statistics, 1356--1378.

\bibitem[{Kock and Callot(2015)}]{kockcallot15}
Kock, A.~B., Callot, L., 2015. Oracle inequalities for high dimensional vector
  autoregressions. Journal of Econometrics 186~(2), 325 -- 344.

\bibitem[{Manzan(2015)}]{manzan15}
Manzan, S., 2015. Forecasting the distribution of economic variables in a
  data-rich environment. Journal of Business \& Economic Statistics 33~(1),
  144--164.

\bibitem[{Meinshausen(2007)}]{meinshausen07}
Meinshausen, N., 2007. Relaxed {Lasso}. Computational statistics and data
  analysis 52~(1), 374--393.

\bibitem[{Meinshausen and B{\"u}hlmann(2006)}]{meinshausenbuhlmann06}
Meinshausen, N., B{\"u}hlmann, P., 2006. High-dimensional graphs and variable
  selection with the {Lasso}. The Annals of Statistics, 1436--1462.

\bibitem[{Meinshausen and Yu(2009)}]{meinshausenyu09}
Meinshausen, N., Yu, B., 2009. {Lasso}-type recovery of sparse representations
  for high-dimensional data. The Annals of Statistics, 246--270.

\bibitem[{Newey and McFadden(1994)}]{neweymcfadden94}
Newey, W.~K., McFadden, D., 1994. Large sample estimation and hypothesis
  testing. Handbook of econometrics 4, 2111--2245.

\bibitem[{Pistoresi et~al.(2011)Pistoresi, Salsano, and
  Ferrari}]{pistoresiall11}
Pistoresi, B., Salsano, F., Ferrari, D., 2011. Political institutions and
  central bank independence revisited. Applied Economics Letters 18~(7),
  679--682.

\bibitem[{Schneider and Wagner(2012)}]{schneiderwagner12}
Schneider, U., Wagner, M., 2012. Catching growth determinants with the adaptive
  lasso. German Economic Review 13~(1), 71--85.

\bibitem[{Schwarz(1978)}]{schwarz78}
Schwarz, G.~E., 1978. Estimating the dimension of a model. Annals of Statistics
  6~(2), 461--464.

\bibitem[{Shao(1997)}]{shao97}
Shao, J., 1997. Asymptotic theory for model selection. Statistica Sinica 7,
  221--242.

\bibitem[{Stone(1974)}]{stone74}
Stone, M., 1974. Cross-validatory choice and assessment of statistical
  predictions. Journal of the Royal Statistical Society, Series B
  (Methodological) 36~(2), 111--147.

\bibitem[{Stone(1977)}]{stone77}
Stone, M., 1977. An asymptotic equivalence of choice of model by
  cross-validation and akaike's criterion. Journal of the Royal Statistical
  Society, Series B (Methodological) 39~(1), 44--47.

\bibitem[{Tibshirani(1996)}]{tibshirani96}
Tibshirani, R., 1996. Regression shrinkage and selection via the {Lasso}.
  Journal of the Royal Statistical Society, Series B (Methodological) 58,
  267--288.

\bibitem[{Tikhonov(1963)}]{tikhonov1963}
Tikhonov, A., 1963. Solution of incorrectly formulated problems and the
  regularization method. In: Soviet Math. Dokl. Vol.~5. pp. 1035--1038.

\bibitem[{Tropp(2004)}]{tropp04}
Tropp, J.~A., 2004. Greed is good: Algorithmic results for sparse
  approximation. Information Theory, IEEE Transactions on 50~(10), 2231--2242.

\bibitem[{Vapnik and Chervonenkis(1971{\natexlab{a}})}]{vc71}
Vapnik, V.~N., Chervonenkis, A.~Y., 1971{\natexlab{a}}. On the uniform
  convergence of relative frequencies of events to their probabilities.
  Theoretical Probability and its Applications 16~(2), 264--280.

\bibitem[{Vapnik and Chervonenkis(1971{\natexlab{b}})}]{vc71b}
Vapnik, V.~N., Chervonenkis, A.~Y., 1971{\natexlab{b}}. Theory of uniform
  convergence of frequencie of appearance of attributes to their probabilities
  and problems of defining optimal solution by empiric data. Avtomatika i
  Telemekhanika~(2), 42--53.

\bibitem[{Vapnik and Chervonenkis(1974)}]{vc74b}
Vapnik, V.~N., Chervonenkis, A.~Y., 1974. On the method of ordered risk
  minimization, {II}. Avtomatika i Telemekhanika~(9), 29--39.

\bibitem[{Varian(2014)}]{varian14}
Varian, H.~R., 2014. Big data: new tricks for econometrics. The Journal of
  Economic Perspectives 28~(2), 3--27.

\bibitem[{Zhang(2010)}]{zhang10}
Zhang, C.-H., 2010. Nearly unbiased variable selection under minimax concave
  penalty. The Annals of Statistics 38, 894--942.

\bibitem[{Zhang and Huang(2008)}]{zhanghuang08}
Zhang, C.-H., Huang, J., 2008. The sparsity and bias of the {Lasso} selection
  in high-dimensional linear regression. The Annals of Statistics 36,
  1567--1594.

\bibitem[{Zhang(2009)}]{zhang09}
Zhang, T., 2009. On the consistency of feature selection using greedy least
  squares regression. Journal of Machine Learning Research 10, 555--568.

\bibitem[{Zhao and Yu(2006)}]{zhaoyu06}
Zhao, P., Yu, B., 2006. On model selection consistency of {Lasso}. The Journal
  of Machine Learning Research 7, 2541--2563.

\bibitem[{Zou(2006)}]{zou06}
Zou, H., 2006. The adaptive {Lasso} and its oracle properties. Journal of the
  American statistical association 101~(476), 1418--1429.

\end{thebibliography}

\newpage{}
\section*{Appendix 1}

\begin{proof}
\textbf{Theorem 1.} Define $b_{test}=\mathrm{argmin}_{b}\;\mathcal{R}_{n_s}\left(b\vert X_s,Y_s\right)$ and $b_{train}=\mathrm{argmin}_{b}\;\mathcal{R}_{n_t}\left(b\vert X_t,Y_t\right)$. The VC inequality (\ref{lem1}) forms an upper bound for the generalization error with probability $1-\eta$, $\forall b$,
\[
\mathcal{R}(b|X,Y) \leqslant \mathcal{R}_{n_t}(b|X_{t},Y_{t})\left(1-\sqrt{\epsilon}\right)^{-1}
\]
where $\mathcal{R}_{n_t}(b|X_{t},Y_{t})$ stands for the training error on $(X_{t},Y_{t})$, $\mathcal{R}(b|X,Y)$ stands for the true generalization error of $b$ and
$\epsilon=(1/n_t)\left\{h\ln\left[\left(n_t/h\right)\right]+h-\ln\left(\eta\right)\right\} $.

Denote $\overline{M}={R}_{n_t}\left(b_{train}\vert X_t,Y_t\right)\left(1-\sqrt{\epsilon}\right)^{-1}$. If we set $\eta=1/n_t$ for $\epsilon$, the VC inequality forms a probabilistic bound for the GE. If $(n_t/h)\rightarrow\infty$, then
\[
\lim_{\widetilde{n}\rightarrow\infty}\epsilon=\lim_{\widetilde{n}\rightarrow\infty}\frac{1}{n_t/h}(\ln[(n_t/h)]+1)
+\lim_{\widetilde{n_t}\rightarrow\infty}\frac{1}{n_t}\ln(n_t)=0.
\]
Thus, the VC inequality is equal to
\[
\lim_{\widetilde{n}\rightarrow\infty}\mathrm{P}
\left\{\left|\overline{M}-\mathcal{R}_{n_t}(b_{train}|X_t,Y_t)\right|\geqslant 1/{n_t}\right\} =0,\quad\forall b_{train}
\]
Given the extremum estimator exists, its loss is finite. Hence, the loss for each data point in the test set $\mathrm{Loss}(y_{i},\widehat{m}(x_{i},b))\in [0,B_i],~\forall i \leqslant n_s$, where $B_i$ is the supremum of $\mathrm{Loss}(y_{i},\widehat{m}(x_{i},b))$. Also, since the extremum estimator converges in the $\mathcal{L}_\infty$ norm,
\[
\lim_{\widetilde{n}\rightarrow\infty}\mathrm{P}
\left\{\sup_{b\in\Lambda}\left|\mathcal{R}_{n_s}(b|X_s,Y_s)-
\mathcal{R}(b|X,Y)\right|\leqslant\varsigma\right\}
=1,\quad\forall\varsigma\geqslant0.
\]
Thus, the upper bound and lower bound of $\vert\mathcal{R}_{n_s}\left(b\vert X_{s},Y_{s}\right)-\mathcal{R}\left(b\vert X,Y\right)\vert$
both converge to 0. Consider the worst case and suppose that $\mathrm{Loss}(y_{i},\widehat{m}(x_{i},b_{train}))$ has heavy tails with the property that, for $1<p\leqslant 2$, $\exists\tau$, such that
\[
\sup_{b\in\Lambda}
\frac{\sqrt[p]{\int[\mathrm{Loss}(y_{i},\widehat{m}(x_{i},b))]^p dF\left(x,y\right)}}
{\int\mathrm{Loss}(y_{i},\widehat{m}(x_{i},b))dF\left(x,y\right)}
\leqslant \tau.
\]
If either tail is sufficiently heavy such that the ratio above is unbounded, VC theory cannot offer a lower bound for the convergence rate or probability computationally. Given this worst case, the Bahr-Esseen inequality
\begin{eqnarray*}
\mathrm{P}\{\vert\mathcal{R}(b_{train}|X,Y)-\mathcal{R}_{n_s}(b_{train}|X_s,Y_s)\vert\leqslant\varsigma\}
& \geqslant 1-2\cdot \frac{\mathbb{E}\left[\mathrm{Loss}(y_{i},\widehat{m}(x_{i},b_{train}))^p\right]}{\varsigma^p\cdot n_s^{p-1}} \\
& \geqslant 1-2\tau^p\cdot \frac{\left(\mathbb{E}\left[\mathrm{Loss}(y_{i},\widehat{m}(x_{i},b_{train}))\right]\right)^p}{\varsigma^p\cdot n_s^{p-1}}
\end{eqnarray*}
holds true for the extremum estimator $b_{train}$. If we define $\varpi= 1-2\tau^p\cdot \left(\mathbb{E}\left[\mathrm{Loss}(y_{i},\widehat{m}(x_{i},b_{train}))\right]\right)^p/(\varsigma^p\cdot n_s^{p-1})$, then
\[
\varsigma = \frac
{\sqrt[p]{2}\cdot\tau\left(\mathbb{E}\left[\mathrm{Loss}(y_{i},\widehat{m}(x_{i},b_{train}))\right]\right)}
{\sqrt[p]{1-\varpi}\cdot n_s^{1-1/p}}
\]
This implies, for any extremum estimator $b_{train}$
\[
\mathrm{P}\{\mathcal{R}_{n_s}(b_{train}|X_{s},Y_{s})
\leqslant\mathcal{R}(b_{train}|X,Y)+\varsigma\} \geqslant \varpi.
\]
The VC inequality holds with probability $1-1/n_t$. For a given $n_s$, we can adapt the probabilistic bound of the empirical process above as follows
\[
\forall\;b_{train}\in\{b_{\lambda}\},
\forall\;\varsigma(1/n_t)=\mathbf{O}(1/n_t)\geqslant 0,
\exists\;N_{t}\in\mathbb{R}^{+}\,\mathrm{s.t.}\;n_t\geqslant N_{t}
\]
\[
\mathcal{R}_{n_s}(b_{train}|X_{s},Y_{s})\leqslant
\varsigma+\overline{M}.
\]
We can relax the bound as follows: $\forall\varsigma\geqslant0,\,\forall\tau_{1}\geqslant0,\,\exists N_{1}\in\mathbb{R}^{+}$ subject to
\[
\mathrm{P}\left\{\mathcal{R}_{n_s}\left(b_{train}|X_{s},Y_{s}\right) \leqslant\frac{\mathcal{R}_{n_t}(b_{train}|X_{t},Y_{t})}{1-\sqrt{\epsilon}}+
\varsigma\right\}
\geqslant \varpi\left(1-\frac{1}{n_t}\right)
\]
Hence, the probabilistic bound $\mathcal{R}_{n_s}(b_{train}|X_{s},Y_{s})\leqslant \overline{M}+\varsigma$ holds with probability at least $\varpi(1-1/n_t)$
\end{proof}

\bigskip{}
\begin{proof}
\textbf{Corollary 1.} Based on Theorem~1, for any extremum estimator $b_{train}$,
\[
\lim_{\widetilde{n}\rightarrow\infty}\mathrm{P}\{\mathcal{R}_{n_s}(b_{train}|X_{s},Y_{s})\leqslant \overline{M}+\varsigma\}=1.
\]
It follows that
\[
\lim_{\widetilde{n}\rightarrow\infty}\mathrm{P}\{\mathcal{R}_{n_s}(b_{test}|X_{s},Y_{s})\leqslant
\mathcal{R}_{n_s}(b_{train}|X_{s},Y_{s}) \leqslant \overline{M}+\varsigma\}=1
\]
Also, since $\varsigma$ could be any small postive value as $\widetilde{n}\rightarrow\infty$ and
\[
\lim_{\widetilde{n}\rightarrow\infty}\left\{\mathcal{R}_{n_s}(b_{test}|X_{s},Y_{s})\right\} =\lim_{\widetilde{n}\rightarrow\infty}\overline{M}
\]
the empirical GE minimizer and structural risk minimizer share the same limit.
\end{proof}

\bigskip{}
\begin{proof}
\textbf{Proposition~1.} Given A1--A4, the true DGP is
\[
y_{i} = x_{i}^{T}\beta + u_{i}, \quad i=1,\ldots,n.
\]
Proving that the true DGP has the highest generalization ability (the lowest GE) is equivalent to proving, in a test set, that
\begin{equation}
\frac{\sum_{i=1}^n\left(y_{i}-x_{i}^{T}\beta\right)^{2}}{n} \leqslant \frac{\sum_{i=1}^n\left(y_{i}-x_{i}^{T}b\right)^{2}}{n},
\label{a1p1}
\end{equation}
which is equivalent to proving that
\begin{align*}
0 & \leqslant \frac{1}{n} \sum_{i=1}^n\left[\left(y_{i}-x_{i}^{T}b\right)^{2}-\left(y_{i}-x_{i}^{T}\beta\right)^{2}\right] \\
\iff
0 & \leqslant \frac{1}{n}
\sum_{i=1}^n\left(y_{i}-x_{i}^{T}b+y_{i}-x_{i}^{T}\beta\right)
\left(y_{i}-x_{i}^{T}b-y_{i}+x_{i}^{T}\beta\right) \\
\iff
0 & \leqslant \frac{1}{n}
\sum_{i=1}^n\left(y_{i}-x_{i}^{T}b+y_{i}-x_{i}^{T}\beta\right)
\left(x_{i}^{T}\beta-x_{i}^{T}b\right).
\end{align*}
Defining $\delta=\beta-b$, it follows,
\begin{align*}
     0 & \leqslant \frac{1}{n}
\sum_{i=1}^n\left(2y_{i}-x_{i}^{T}b-x_{i}^{T}\beta\right)
\left(x_{i}^{T}\delta\right) \\
\iff 0 & \leqslant \frac{1}{n} \sum_{i=1}^n\left(2y_{i}-x_{i}^{T}\beta+x_{i}^{T}\beta-x_{i}^{T}b-x_{i}^{T}\beta\right)
\left(x_{i}^{T}\delta\right) \\
\iff 0 & \leqslant \frac{1}{n}
\sum_{i=1}^n\left(2y_{i}-2x_{i}^{T}\beta+x_{i}^{T}\delta\right)
\left(x_{i}^{T}\delta\right) \\
\iff 0 & \leqslant \frac{1}{n}
\sum_{i=1}^n\left(2u_{i}+x_{i}^{T}\delta\right)\left(x_{i}^{T}\delta\right)
\end{align*}
Hence, proving (\ref{a1p1}) is equivalent to proving
\[
0 \leqslant \frac{1}{n} \sum_{i=1}^n\left(2u_{i}+x_{i}^{T}\delta\right)\left(x_{i}^{T}\delta\right)
\]
Since $\mathbb{E}(X^T u)=\mathbf{0}$ (A2), it follows that
\[
\frac{1}{n}\sum_{i=1}^n u_{i}\cdot x_{i}\overset{\mathbf{P}}{\rightarrow}\mathbf{0}
\iff
\frac{1}{n}\sum_{i=1}^n\left(u_{i}\cdot x_{i}^{T}\right)\beta\overset{\mathbf{P}}{\rightarrow}0
\quad\mbox{and}\quad
\frac{1}{n}\stackrel[i=1]{n}{\sum}\left(u_{i}\cdot x_{i}^{T}\right)b\rightarrow 0
\]
Hence, asymptotically
\[
\frac{1}{n}\sum_{i=1}^n\left(2u_{i}+x_{i}^{T}\delta\right)\left(x_{i}^{T}\delta\right)=
\frac{1}{n}\sum_{i=1}^n 2\delta u_{i}x_{i}^{T}+
\frac{1}{n}\sum_{i=1}^n\left(x_{i}^{T}\delta\right)^{2}
\overset{\mathbf{P}}{\rightarrow}\mathbb{E}\left(x_{i}^{T}\delta\right)^{2}\geqslant 0
\]
\end{proof}

\bigskip{}
\begin{proof}
\textbf{Theorem~2.} In the proof of Theorem~1 we defined $b_{train}=\mathrm{argmin}_{b}\;\mathcal{R}_{n_t}\left(b\vert X_t,Y_t\right)$, meaning that $b_{train}$ is the extremum estimator without penalty on any training set. We also have $\overline{M}={R}_{n_t}\left(b_{train}\vert X_t,Y_t\right)\left(1-\sqrt{\epsilon}\right)^{-1}$. Theorem~1 shows the following bound holds $\forall b\in\Lambda $
\[
\mathcal{R}_{n_s}(b|X_{s},Y_{s}) \leqslant
\mathcal{R}_{n_t}(b|X_{t},Y_{t})\left(1-\sqrt{\epsilon}\right)^{-1}+
\varsigma
\]
with probability at least $(1-1/n_t)\varpi$. Also, among all the $b\in\{ b_{\lambda}\}$, $b_{Lasso}$ has the lowest GE on the test set,
\[
\mathcal{R}_{n_s}(b_{Lasso}|X_{s},Y_{s})\leqslant\mathcal{R}_{n_s}(b|X_{s},Y_{s})
\]
we have
\[
\frac{1}{n_s}\left\Vert Y_{s}-X_{s}b_{Lasso}\right\Vert _{2}^{2} \leqslant
\frac{1}{n_t}\left\Vert Y_{t}-X_{t}b_{train}\right\Vert_{2}^{2}\left(1-\sqrt{\epsilon}\right)^{-1}+\varsigma
\]
By defining $\Delta=b_{train}-b_{Lasso}$, $Y_{t}-X_{t}b_{train}=e_{t}$ and $Y_{s}-X_{s}b_{train}=e_{s}$,
\begin{align*}
\frac{1}{n_s}\left\Vert Y_{s}-X_{s}b_{Lasso}\right\Vert_{2}^{2}
& =\frac{1}{n_s}\left\Vert Y_{s}-X_{s}b_{train}+X_{s}\Delta\right\Vert_{2}^{2}\\
& =\frac{1}{n_s}\left\Vert e_{s}+X_{s}\Delta\right\Vert_{2}^{2}\\
& =\frac{1}{n_s}\left(e_{s}+X_{s}\Delta\right)^{T}\left(e_{s}+X_{s}\Delta\right)\\
& =\frac{1}{n_s}\left(\left\Vert e_{s}\right\Vert_{2}^{2}+2e_{s}^{T}X_{s}\Delta+
\Delta^{T}X_{s}^{T}X_{s}\Delta\right)
\end{align*}
Hence,
\[
\frac{1}{n_s}\left\Vert Y_{s}-X_{s}b_{Lasso}\right\Vert_{2}^{2} \leqslant
\frac{1}{n_t}\left\Vert Y_{t}-X_{t}b_{train}\right\Vert_{2}^{2}\left(1-\sqrt{\epsilon}\right)^{-1}+\varsigma
\]
implies
\[
\frac{1}{n_s}\left\Vert e_{s}\right\Vert_{2}^{2}+\frac{2}{n_s}e_{s}^{T}X_{s}\Delta+
\frac{1}{n_s}\Delta^{T}X_{s}^{T}X_{s}\Delta \leqslant
\frac{\frac{1}{n_t}\left\Vert e_{t}\right\Vert_{2}^{2}}{1-\sqrt{\epsilon}}+
\varsigma.
\]
It follows that
\[
\frac{1}{n_s}\left\Vert X_{s}\Delta\right\Vert_{2}^{2} \leqslant
\left(\frac{1}{n_t}\frac{\left\Vert e_{t}\right\Vert_{2}^{2}}{1-\sqrt{\epsilon}}-
\frac{1}{n_s}\left\Vert e_{s}\right\Vert_{2}^{2}\right)-
\frac{2}{n_s}e_{s}^{T}X_{s}\Delta+\varsigma.
\]
By the Holder inequality,
\[
-e_{s}^{T}X_{s}\Delta\leqslant\vert e_{s}^{T}X_{s}\Delta\vert\leqslant\left\Vert e_{s}^{T}X_{s}\right\Vert_{\infty}\left\Vert\Delta\right\Vert_{1}.
\]
It follows that
\[
\frac{1}{n_s}\left\Vert X_{s}\Delta\right\Vert_{2}^{2} \leqslant
\left(\frac{1}{n_t}\frac{\left\Vert e_{t}\right\Vert_{2}^{2}}{1-\sqrt{\epsilon}}-
\frac{1}{n_s}\left\Vert e_{s}\right\Vert_{2}^{2}\right)+
\frac{2}{n_s}\left\Vert e_{s}^{T}X_{s}\right\Vert_{\infty}\left\Vert\Delta\right\Vert _{1}+\varsigma.
\]
Also, since $\left\Vert b_{Lasso}\right\Vert_{1}\leqslant\left\Vert b_{train}\right\Vert_{1}$
\begin{align*}
\left\Vert \Delta\right\Vert _{1} & = \left\Vert b_{train}-b_{Lasso}\right\Vert_{1}\\
& \leqslant  \left\Vert b_{Lasso}\right\Vert_{1}+\left\Vert b_{train}\right\Vert_{1}\\
& \leqslant 2\left\Vert b_{train}\right\Vert_{1}
\end{align*}
As a result, we have
\begin{equation}
\frac{1}{n_s}\left\Vert X_{s}\Delta\right\Vert _{2}^{2} \leqslant
\left(\frac{1}{n_t}\frac{\left\Vert e_{t}\right\Vert_{2}^{2}}{1-\sqrt{\epsilon}}-
\frac{1}{n_s}\left\Vert e_{s}\right\Vert_{2}^{2}\right)+
\frac{4}{n_s}\left\Vert e_{s}^{T}X_{s}\right\Vert_{\infty}\left\Vert b_{train}\right\Vert_{1}+\varsigma
\label{a1t2}
\end{equation}
where we denote $\widehat{y}_s$ to be the extremum estimator prediction on the test set calculated on the training set and $\widehat{y}_s^{Lasso}$ to be the Lasso prediction on the test set. It follows that (\ref{a1t2}) is the bound for
$\mathbb{E}\left(\left\Vert \widehat{y_{s}}-\widehat{y}_{s}^{Lasso}\right\Vert _{2}^{2}\right)$,
the expected difference between Lasso prediction and extremum estimator prediction on the test set. The bound holds with probability $(1-1/n_t)\varpi$.
\end{proof}

\bigskip{}
\begin{proof}
\textbf{Corollary~2.} We need to prove that the VC inequality and SRM also hold for Lasso with cross-validation. If Lasso is implemented by $K$-fold cross-validation, the sample is partitioned into $K$ equal-sized folds. If $K=2$, the theoretical result from $K$-fold cross-validation is identical to Theorem~2. Thus, we only discuss the case of $K\geqslant 3$ here.

For $K\geqslant 3$, we have $K$ different test sets for $\lambda$-tuning and $K$ different training sets for estimation. Denote the $q$\textsuperscript{th} training set as $\left(X_t^q,Y_t^q\right)$, the $q$\textsuperscript{th} test set as $\left(X_s^q,Y_s^q\right)$, the extremum estimator estimated from the $k$\textsuperscript{th} training set as $b_{train}^k$, the sample size for each test set as $n_s$ and the sample size for each training set as $n_t$. Based on Theorem 1, for each test set, the following bound holds for $k\;\mbox{and}\;q\in\left[1,K\right]$ with probability at least $(1-1/n_t)\varpi_k$
\[
\mathcal{R}_{n_s}(b_{train}^k|X_s^q,Y_s^q) \leqslant
\mathcal{R}_{n_t}(b_{train}^k|X_t^q,Y_t^q)\left(1-\sqrt{\epsilon_k}\right)^{-1}+
\varsigma_{k}.
\]
Hence,
\begin{align*}
\frac{1}{K}\sum_{q=1}^K\mathcal{R}_{n_s}(b_{train}^k|X_s^q,Y_s^q) & \leqslant
\mathcal{R}_{n_s}(\thickbar{b}_{train}|X_s^{q^*},Y_s^{q^*}) \\
&\leqslant \mathcal{R}_{n_t}(\thickbar{b}_{train}|X_t^{q^*},Y_t^{q^*})\left(1-\sqrt{\thickbar{\epsilon}}\right)^{-1}
+\thickbar{\varsigma}
\end{align*}
Since $b_{Lasso}$ minimizes $(1/K)\sum_{q=1}^K\mathcal{R}_{n_s}(b|X_s^q,Y_s^q)$,
\[
\frac{1}{K}\sum_{q=1}^K\mathcal{R}_{n_s}(b_{Lasso}|X_s^q,Y_s^q) \leqslant
\frac{1}{K}\sum_{q=1}^K\mathcal{R}_{n_s}(b_{train}^k|X_s^q,Y_s^q),
\quad\forall k\in[1,K]
\]
It follows that
\[
\frac{1}{K}\sum_{q=1}^K\mathcal{R}_{n_s}(b_{Lasso}|X_s^q,Y_s^q) \leqslant \mathcal{R}_{n_t}(\thickbar{b}_{train}|X_t^{q^*},Y_t^{q^*})
\left(1-\sqrt{\thickbar{\epsilon}}\right)^{-1}+\thickbar{\varsigma}.
\]
Denote $\mathcal{R}_{n_t}(b_{train}^{k^*}|X_t^{q^*},Y_t)$ by $\overline{e_t}$ and $\mathcal{R}_{n_s}(b_{train}^{k^*}|X_s^{q^*},Y_s^{q^*})$ by $\overline{e_s}$. The above equation is equivalent to
\[
\frac{1}{K}\sum_{q=1}^K \left(\frac{1}{n_s}\Vert Y_s^q-X_s^qb_{Lasso}\Vert_2^2\right) \leqslant
\frac{\Vert \overline{e_t}\Vert_2^2}{n_t}\frac{1}{1-\sqrt{\thickbar{\epsilon}}}+\thickbar{\varsigma}.
\]
By defining $\Delta=\thickbar{b}_{train} - b_{Lasso}$ and $\overline{e_s^q} = Y_s^q-X_s^q\thickbar{b}_{train}$ we have
\begin{align*}
\frac{1}{n_s}\left\Vert Y_s^q-X_s^qb_{Lasso}\right\Vert_{2}^{2}
& =\frac{1}{n_s}\left\Vert Y_s^q-X_s^q\thickbar{b}_{train}+X_s^q\Delta\right\Vert_{2}^{2}\\
& =\frac{1}{n_s}\left\Vert \overline{e_s^q}+X_s^q\Delta\right\Vert_{2}^{2}\\
& =\frac{1}{n_s}\left(\overline{e_s^q}+X_s^q\Delta\right)^{T}\left(\overline{e_s^q}+X_s^q\Delta\right)\\
& =\frac{1}{n_s}\left(\left\Vert \overline{e_s^q}\right\Vert_{2}^{2}+2\left(\overline{e_s^q}\right)^{T}X_s^q\Delta+
\Delta^{T}\left(X_s^q\right)^{T}X_s^q\Delta\right).
\end{align*}
Hence,
\[
\frac{1}{K}\sum_{q=1}^K \left(\frac{1}{n_s}\left\Vert Y_s^q-X_s^qb_{Lasso}\right\Vert_{2}^{2}\right)
\leqslant
\frac{1}{n_t}\left\Vert Y_t^{q}-X_t^{q}\thickbar{b}_{train}\right\Vert_{2}^{2}\left(1-\sqrt{\thickbar{\epsilon}}\right)^{-1}
+\thickbar{\varsigma}
\]
implies
\[
 \frac{1}{K}\sum_{q=1}^K\frac{1}{n_s}\left\Vert \overline{e_s^q}\right\Vert_{2}^{2}
+\frac{1}{K}\sum_{q=1}^K\frac{2}{n_s}\left(\overline{e_s^q}\right)^{T}X_{s}\Delta
+\frac{1}{K}\sum_{q=1}^K\frac{1}{n_s}\Delta^{T}\left(X_s^q\right)^{T}\left(X_s^q\right)\Delta
 \leqslant
\frac{\frac{1}{n_t}\left\Vert\overline{e_t}\right\Vert_{2}^{2}}{1-\sqrt{\thickbar{\epsilon}}}
+\thickbar{\varsigma}.
\]
It follows that
\[
 \frac{1}{K}\sum_{q=1}^K\frac{1}{n_s}\left\Vert X_s^q\Delta\right\Vert_{2}^{2}
 \leqslant
\frac{1}{n_t}\frac{\left\Vert \overline{e_t}\right\Vert_{2}^{2}}{1-\sqrt{\thickbar{\epsilon}}}
-\frac{1}{K}\sum_{q=1}^K\frac{\left\Vert \overline{e_s^q}\right\Vert_{2}^{2}}{n_s}
-\frac{1}{K}\sum_{q=1}^K\frac{2}{n_s}\left(\overline{e_s^q}\right)^{T}X_{s}^q\Delta
+\thickbar{\varsigma}.
\]
By the Holder inequality,
\[
-1\cdot \left(\overline{e_s^q}\right)^{T}X_s^q\Delta\leqslant\vert \left(\overline{e_s^q}\right)^{T}X_s^q\Delta\vert\leqslant\left\Vert \left(\overline{e_s^q}\right)^{T}X_s^q\right\Vert_{\infty}\left\Vert\Delta\right\Vert_{1}.
\]
It follows that
\[
 \frac{1}{K}\sum_{q=1}^K\frac{1}{n_s}\left\Vert X_s^q\Delta\right\Vert_{2}^{2}
\leqslant
\left\vert\frac{1}{n_t}\frac{\left\Vert \overline{e_{t}}\right\Vert_{2}^{2}}{1-\sqrt{\thickbar{\epsilon}}}
-\frac{1}{K}\sum_{q=1}^K\frac{\left\Vert \overline{e_s^q}\right\Vert_{2}^{2}}{n_s}\right\vert
+\frac{1}{K}\sum_{q=1}^K\frac{2}{n_s}\Vert\left(\overline{e_s^q}\right)^{T}X_{s}^q\Vert_{\infty}
\Vert\Delta\Vert_{1}
+\thickbar{\varsigma}.
\]
Also, since $\left\Vert b_{Lasso}\right\Vert_{1}\leqslant\left\Vert \thickbar{b}_{train}\right\Vert_{1}$
\begin{align*}
\left\Vert \Delta\right\Vert _{1} & = \left\Vert \thickbar{b}_{train}-b_{Lasso}\right\Vert_{1}\\
& \leqslant  \left\Vert b_{Lasso}\right\Vert_{1}+\left\Vert \thickbar{b}_{train}\right\Vert_{1}\\
& \leqslant 2\left\Vert \thickbar{b}_{train}\right\Vert_{1}
\end{align*}
Therefore, we have
\[
\frac{1}{K}\sum_{q=1}^K \frac{1}{n_s}\left\Vert X_s^q\Delta\right\Vert _{2}^{2}
 \leqslant
\left\vert\frac{1}{n_t}\frac{\left\Vert \overline{e_t}\right\Vert_{2}^{2}}{1-\sqrt{\thickbar{\epsilon}}}
-\frac{1}{K}\sum_{q=1}^K\frac{1}{n_s}\left\Vert \overline{e_s^q}\right\Vert_{2}^{2}\right\vert
+\frac{1}{K}\sum_{q=1}^K\frac{4}{n_s}\left\Vert \left(\overline{e_s^q}\right)^TX_s^q\right\Vert_{\infty}\left\Vert \thickbar{b}_{train}\right\Vert_{1}
+\thickbar{\varsigma}
\]
This formula is the bound for
$\mathbb{E}_k\left[\mathbb{E}_{\left(X_s^k,Y_s^k\right)}\left(\left\Vert \widehat{y_{s}}-\widehat{y}_{s}^{Lasso}\right\Vert _{2}^{2}\right)\right]$,
the iterated expected difference between Lasso prediction and extremum prediction on any $\left(X_s^k,Y_s^k\right)$. The bound holds with probability $(1-1/n_t)\thickbar{\varpi}$.

\end{proof}

\bigskip{}
\begin{proof}
\textbf{Theorem~3.} (Consistency when $n\geqslant p$.) Under the \citet{neweymcfadden94} condition, the extremum estimators is consistent. If $n\geqslant p$, the extremum estimator $b_{train}$ is simply the OLS estimator. We prove that Lasso tuned by validation is consistent for $n\geqslant p$.

As long as $(\widetilde{n}/p)\rightarrow\infty$, $\mathcal{R}_{n_t}(b_{train}|X_{t},Y_{t})\overset{\mathbf{P}}{\rightarrow}\inf_b\mathcal{\,R}(b|X,Y)$
and $\mathcal{R}_{n_s}(b_{train}|X_{s},Y_{s})\overset{\mathbf{P}}{\rightarrow}\inf_b\mathcal{R}(b|X,Y)$,
which means $(1/n_t)\left\Vert e_{t}\right\Vert_{2}^{2}$ and $(1/n_s)\left\Vert e_{s}\right\Vert_{2}^{2}$ all converge to the same limit. As a result,
\[
\frac{(1/n_t)\left\Vert e_{t}\right\Vert_{2}^{2}}
{{(1/n_s)}\left\Vert e_{s}\right\Vert_{2}^{2}}
\overset{\mathbf{P}} {\longrightarrow} 1.
\]
Also $(4/n_s)\left\Vert e_{s}^{T}X_{s}\right\Vert_{\infty}\overset{\mathbf{P}}{\rightarrow}0$,
$\left\Vert b_{train}\right\Vert_{1}\rightarrow\left\Vert\beta\right\Vert_{1}$ and $\epsilon\rightarrow0$. Also $\widehat{y}\overset{\mathbf{P}}{\rightarrow}X\beta$ if $(n/p)\rightarrow\infty$. Hence $Xb_{Lasso}\overset{\mathcal{L}_{2}}{\rightarrow}X\beta$.

For OLS, $(1/n)\left\Vert X_{s}\Delta\right\Vert _{2}^{2}\geqslant\rho\left\Vert \Delta\right\Vert _{2}^{2}$, where $\rho$ is the minimal eigenvalue for $X^{T}X$. Hence,
\begin{align*}
\rho\left\Vert \Delta\right\Vert _{2}^{2} &
\leqslant
\frac{1}{n_s}\left\Vert X_{s}\Delta\right\Vert _{2}^{2} \\
& \leqslant
\left\vert\frac{1}{n_t}\frac{\left\Vert e_{t}\right\Vert_{2}^{2}}{(1-\sqrt{\epsilon})}
-\frac{1}{n_s}\left\Vert e_{s}\right\Vert_{2}^{2}\right\vert
+\frac{4}{n_s}\left\Vert e_{s}^{T}X_{s}\right\Vert_{\infty}\left\Vert b_{train}\right\Vert_{1}+\varsigma.
\end{align*}
It follows that
\[
\rho\left\Vert \Delta\right\Vert _{2}^{2} \leqslant
\left\vert\frac{1}{n_t}\frac{\left\Vert e_{t}\right\Vert_{2}^{2}}{(1-\sqrt{\epsilon})}
-\frac{1}{n_s}\left\Vert e_{s}\right\Vert_{2}^{2}\right\vert
+\frac{4}{n_s}\left\Vert e_{s}^{T}X_{s}\right\Vert_{\infty}\left\Vert b_{train}\right\Vert_{1}+\varsigma.
\]
By the Minkowski inequality, the above can be simplified to
\[
\left\Vert b_{train}-b_{Lasso}\right\Vert _{2} \leqslant
\sqrt{\left|\frac{1}{\rho n_t}\frac{\left\Vert e_{t}\right\Vert_{2}^{2}}{\left(1-\sqrt{\epsilon}\right)}
-\frac{1}{\rho n_s}\left\Vert e_{s}\right\Vert _{2}^{2}\right|}
+\sqrt{\frac{4}{\rho n_s}\left\Vert e_{s}^{T}X_{s}\right\Vert_{\infty}
\left\Vert b_{train}\right\Vert _{1}}\\
+\left(\frac{\varsigma}{\rho}\right)^{\frac{1}{2}}
\]
Thus, the extremum estimator and the Lasso estimator asymptotically converge to $\beta$.
\end{proof}

\bigskip{}
\begin{proof}
\textbf{Corollary~3.} (Consistency when $n\geqslant p$.) If $n\geqslant p$, extremum estimation $\thickbar{b}_{train}$ is the OLS estimator for the `worst case'. We prove that Lasso tuned by cross-validation is consistent for $n\geqslant p$.

For OLS, $(1/n_s)\left\Vert X_s^q\Delta\right\Vert _{2}^{2}\geqslant\rho_q\left\Vert \Delta\right\Vert_{2}^{2}$, where $\rho_q$ is the minimal eigenvalue for $\left(X_s^q\right)^{T}\left(X_s^q\right)$. For cross-validated Lasso,
\[
\frac{1}{K}\sum_{q=1}^K \frac{1}{n_s}\left\Vert X_s^q\Delta\right\Vert _{2}^{2} \leqslant
\left\vert\frac{1}{n_t}\frac{\left\Vert\overline{e_t}\right\Vert_{2}^{2}}{(1-\sqrt{\thickbar{\epsilon}})}
-\frac{1}{K}\sum_{q=1}^K\frac{1}{n_s}\left\Vert\overline{e_s^q}\right\Vert_{2}^{2}\right\vert \\
+\frac{1}{K}\sum_{q=1}^K\frac{4}{n_s}\left\Vert\left(\overline{e_s^q}\right)^TX_s^q\right\Vert_{\infty}
\left\Vert \thickbar{b}_{train}\right\Vert_{1}
+\thickbar{\varsigma}
\]
implies that
\[
\frac{1}{K}\sum_{q=1}^K \rho_q\left\Vert\Delta\right\Vert _{2}^{2} \leqslant
\left\vert\frac{1}{n_t}\frac{\left\Vert \overline{e}_t\right\Vert_{2}^{2}}{(1-\sqrt{\thickbar{\epsilon}})}
-\frac{1}{K}\sum_{q=1}^K\frac{1}{n_s}\left\Vert\overline{e_s^q}\right\Vert_{2}^{2}\right\vert
+\frac{1}{K}\sum_{q=1}^K\frac{4}{n_s}\left\Vert\left(\overline{e_s^q}\right)^TX_s^q\right\Vert_{\infty}
\left\Vert \thickbar{b}_{train}\right\Vert_{1}
+\thickbar{\varsigma}.
\]
Denoting $\min_q \rho_q$ by $\rho^*$,
\[
\frac{\rho^*}{K}\sum_{q=1}^K \left\Vert\Delta\right\Vert _{2}^{2}
\leqslant
\frac{1}{K}\sum_{q=1}^K \rho_q\left\Vert\Delta\right\Vert _{2}^{2}.
\]
Hence,
\[
\frac{\rho^*}{K}\sum_{q=1}^K \left\Vert\Delta\right\Vert _{2}^{2} \leqslant
\left\vert\frac{1}{n_t}\frac{\left\Vert \overline{e_t}\right\Vert_{2}^{2}}{(1-\sqrt{\thickbar{\epsilon}})}
-\frac{1}{K}\sum_{q=1}^K\frac{1}{n_s}\left\Vert \overline{e_s^q}\right\Vert_{2}^{2}\right\vert
+\frac{1}{K}\sum_{q=1}^K\frac{4}{n_s}\left\Vert\left(\overline{e_s^q}\right)^TX_s^q
\right\Vert_{\infty}\left\Vert \thickbar{b}_{train}\right\Vert_{1}
+\thickbar{\varsigma}
\]
implies
\begin{align}
\frac{1}{K}\sum_{q=1}^K \left\Vert \thickbar{b}_{train} - b_{Lasso}\right\Vert _{2}^{2} & \leqslant
\left\vert\frac{1}{n_t\cdot\rho^*}\frac{\left\Vert\overline{e_t}\right\Vert_{2}^{2}}{(1-\sqrt{\thickbar{\epsilon}})}
-\frac{1}{K}\sum_{q=1}^K\frac{1}{n_s\cdot\rho^*}\left\Vert\overline{e_s^q}\right\Vert_{2}^{2}\right\vert \nonumber \\
&
+\frac{1}{K}\sum_{q=1}^K\frac{4}{n_s\cdot\rho^*}\left\Vert\left(\overline{e_s^q}\right)^TX_s^q\right\Vert_{\infty}
\left\Vert \thickbar{b}_{train}\right\Vert_{1}+\frac{\thickbar{\varsigma}}{\rho^*}.
\label{c3}
\end{align}
The equation above is the bound for
\[
\mathbb{E}_k\left[\mathbb{E}_{\left(X_s^k,Y_s^k\right)}
\left[\left\Vert \thickbar{b}_{train} - b_{Lasso}\right\Vert _{2}^{2}\right]\right].
\]
As $n\rightarrow\infty$, the RHS of (\ref{c3}) converges to zero and $\thickbar{b}_{train} \overset{\mathcal{L}_2}{\rightarrow} b_{Lasso}$. Since $\thickbar{b}_{train}$ converges to $\beta$ in $\mathcal{L}_2$, as guaranteed by the asymptotic property of OLS, $b_{Lasso}$ also converges to $\beta$ in $\mathcal{L}_2$
\end{proof}

\bigskip{}
\begin{proof}
\textbf{Theorem 4.} (Consistency when $n < p$.) In this proof we show that Lasso and FSR both converge to the true DGP if Lasso is tuned by validation. For regressions where $n < p$, the OLS estimator is not feasible because $X^TX$ is not of full rank and the traditional strong convexity condition fails. As a result, $(1/n)\left\Vert X_{s}\Delta\right\Vert _{2}^{2}\geqslant\rho\left\Vert \Delta\right\Vert _{2}^{2}$ may not hold for all $\Delta$, the extremum estimator may not converge to $\beta$, and the consistency result established in Theorem~3 may not be valid.

To solve this problem, we import the restricted eigenvalue condition from \citet{bickeletal09} and \citet{meinshausenyu09}.\footnote{\citet{meinshausenyu09} develop a version of the restricted eigenvalue condition, which they call the sparse eigenvalue condition.} The restricted eigenvalue condition assumes that $(1/n)\left\Vert X_{s}\Delta\right\Vert_{2}^{2}\geqslant \widetilde{\rho}\left\Vert \Delta\right\Vert_{2}^{2}$ still holds for all $b \in \left\{b_{\lambda}\right\}$ \citep{bickeletal09} and FSR estimators $b_{train}$ \citep{zhang09}. Also, in this scenario, the extremum estimator $\min_{b}(1/n_t)\Vert Y_t-X_tb\Vert_2^2$ can be implemented by forward selection of at most $n$ variables that minimize the training error. As shown by \citet{tropp04} and \citet{zhang09}, forward selection regression is consistent under the restricted eigenvalue condition.

As long as $n\rightarrow\infty$, $\mathcal{R}_{emp}(b_{train}|X_{t}^{n},Y_{t}^{n})\overset{\mathbf{P}}
{\rightarrow}\inf_b\mathcal{\,R}(b|X,Y)$ and $\mathcal{R}_{emp}(b_{train}|X_{s}^{n},Y_{s}^{n})
\overset{\mathbf{P}}{\rightarrow}\inf_b\mathcal{R}(b|X,Y)$, which means
$(1/n)\left\Vert e_{t}\right\Vert_{2}^{2}$ and $(1/n)\left\Vert e_{s}\right\Vert_{2}^{2}$
all converge to the same limit. Thus,
$(1/n)\left\Vert e_{t}\right\Vert_{2}^{2}-(1/n)\left\Vert e_{s}\right\Vert_{2}^{2}\overset{\mathbf{P}}{\rightarrow}0$.
Also $(4/n)\left\Vert e_{s}^{T}X_{s}\right\Vert_{\infty}\overset{\mathbf{P}}{\rightarrow}0$, $\left\Vert b_{train}\right\Vert _{1}\overset{\mathbf{P}}{\rightarrow}\left\Vert \beta\right\Vert _{1}$ and $\epsilon\rightarrow0$. Also $\widehat{y}^{*}\overset{\mathbf{P}}{\rightarrow}X\beta$ if $(n/p)\rightarrow\infty$. Hence, $Xb_{Lasso}\overset{\mathcal{L}_{2}}{\rightarrow}X\beta$.

For OLS,
$(1/n)\left\Vert X_{s}\Delta\right\Vert_{2}^{2}\geqslant\widetilde{\rho}\left\Vert\Delta\right\Vert_{2}^{2}$,
where $\rho$ is the minimum restricted eigenvalue for $X^{T}X$. Similar to Theorem 3, equation (2) in the proof of Theorem 2 can be simplified to
\[
\left\Vert b_{train}-b_{Lasso}\right\Vert_{2} \leqslant
\sqrt{\left|\frac{1}{\widetilde{\rho} n_t}
\frac{\left\Vert e_{t}\right\Vert_{2}^{2}}{\left(1-\sqrt{\epsilon}\right)}
-\frac{1}{\widetilde{\rho}n_s}\left\Vert e_{s}\right\Vert _{2}^{2}\right|}
+\sqrt{\frac{4}{\widetilde{\rho} n_s}\left\Vert e_{s}^{T}X_{s}\right\Vert_{\infty}
\left\Vert b_{train}\right\Vert_{1}}
+\left(\frac{\varsigma}{\widetilde{\rho}}\right)^{\frac{1}{2}}.
\]
Since \citet{tropp04} and \citet{zhang09} prove that forward selection regression is consistent, it follows that the extremum estimator and Lasso estimator asymptotically converge to $\beta$.
\end{proof}

\bigskip{}
\begin{proof}
\textbf{Corollary~4.} (Consistency when $n<p$.) In this proof we show that, under cross-validation, a very similar bound to Theorem~4 holds for Lasso as well.

As above, denoting the $\min_q\widetilde{\rho}_q$ by $\widetilde{\rho}^*$,
\[
\frac{\widetilde{\rho}^*}{K}\sum_{q=1}^K \left\Vert\Delta\right\Vert _{2}^{2}
\leqslant
\frac{1}{K}\sum_{q=1}^K \widetilde{\rho}_q\left\Vert\Delta\right\Vert _{2}^{2}.
\]
Hence,
\[
\frac{\widetilde{\rho}^*}{K}\sum_{q=1}^K \left\Vert\Delta\right\Vert _{2}^{2} \leqslant
\left\vert\frac{1}{n_t}\frac{\left\Vert \overline{e_t}\right\Vert_{2}^{2}}{1-\sqrt{\thickbar{\epsilon}}}
-\frac{1}{K}\sum_{q=1}^K\frac{1}{n_s}\left\Vert\overline{e_s^q}\right\Vert_{2}^{2}\right\vert
+\frac{1}{K}\sum_{q=1}^K\frac{4}{n_s}\left\Vert\left(\overline{e_s^q}\right)^TX_s^q\right\Vert_{\infty}
\left\Vert \thickbar{b}_{train}\right\Vert_{1}
+\thickbar{\varsigma}
\]
implies
\begin{align}
\frac{1}{K}\sum_{q=1}^K \left\Vert \thickbar{b}_{train} - b_{Lasso}\right\Vert _{2}^{2} &
\leqslant
\left\vert\frac{1}{n_t\cdot\widetilde{\rho}^*}\frac{\left\Vert \overline{e_t}\right\Vert_{2}^{2}}{1-\sqrt{\thickbar{\epsilon}}}
-\frac{1}{K}\sum_{q=1}^K\frac{1}{n_s\cdot\widetilde{\rho}^*}\left\Vert \overline{e_s^q}\right\Vert_{2}^{2}\right\vert \nonumber \\
& +\frac{1}{K}\sum_{q=1}^K\frac{4}{n_s\cdot\widetilde{\rho}^*}\left\Vert \left(\overline{e_s^q}\right)^TX_s^q\right\Vert_{\infty}\left\Vert \thickbar{b}_{train}\right\Vert_{1}
+\frac{\thickbar{\varsigma}}{\widetilde{\rho}^*}.
\label{c4}
\end{align}
The equation above is the bound for
\[
\mathbb{E}_k\left[\mathbb{E}_{\left(X_s^k,Y_s^k\right)}\left[\left\Vert \thickbar{b}_{train}
-b_{Lasso}\right\Vert_{2}^{2}\right]\right].
\]
As $n\rightarrow\infty$, the RHS of (\ref{c4}) converges to zero and $\thickbar{b}_{train} \overset{\mathcal{L}_2}{\rightarrow} b_{Lasso}$. Since $\thickbar{b}_{train}$ converges to $\beta$ in $\mathcal{L}_2$ \citep{tropp04,zhang09}, $b_{Lasso}$ also converges to $\beta$ in $\mathcal{L}_2$
\end{proof}

\newpage
\section*{Appendix 2}

\begin{table}[h]
\caption*{\textbf{Forward selection regression algorithm}}

\begin{tabular}{lp{137mm}}
\toprule
1. & Standardize $Y$ and the variables $X_{j}$, $j=1,\ldots,p$ \\
2. & Start the regression from $Y = u$ \\
3. & Add the variable having the largest correlation with $u$ into the regression and estimate $Y = Xb + u$ \\
4. & Repeat 3, one variable at a time, until the maximum correlation between $u$ and the most recent variable added to the model is less than some preset value.\\
\bottomrule
\end{tabular}
\end{table}

\bigskip
\begin{table}[h]
\caption*{\textbf{CV-Lasso algorithm}}

\begin{tabular}{lp{137mm}}
\toprule
1. & Set $\lambda=0$ \\
2. & by using \textit{k}-fold cross-validation, divide the original sample into a training set $T$ and a test set $S$ \\
3. & Compute the Lasso estimator $b_{\lambda}$ on $T$ and calculate the GE of $Xb_{\lambda}$ on $S$ \\
4. & Increase $\lambda$ by a preset step size and repeat 2 and 3 until $b_{\lambda}=\mathbf{0}$ \\
5. & Pick the $b_{\lambda}$ that minimizes the GE and denote it $b_{Lasso}$ \\
\bottomrule
\end{tabular}
\end{table}

\newpage
\section*{Appendix 3}

\begin{figure}[ht]
\centering
\subfloat[Lasso estimates]
{\includegraphics[width=0.35\paperwidth]{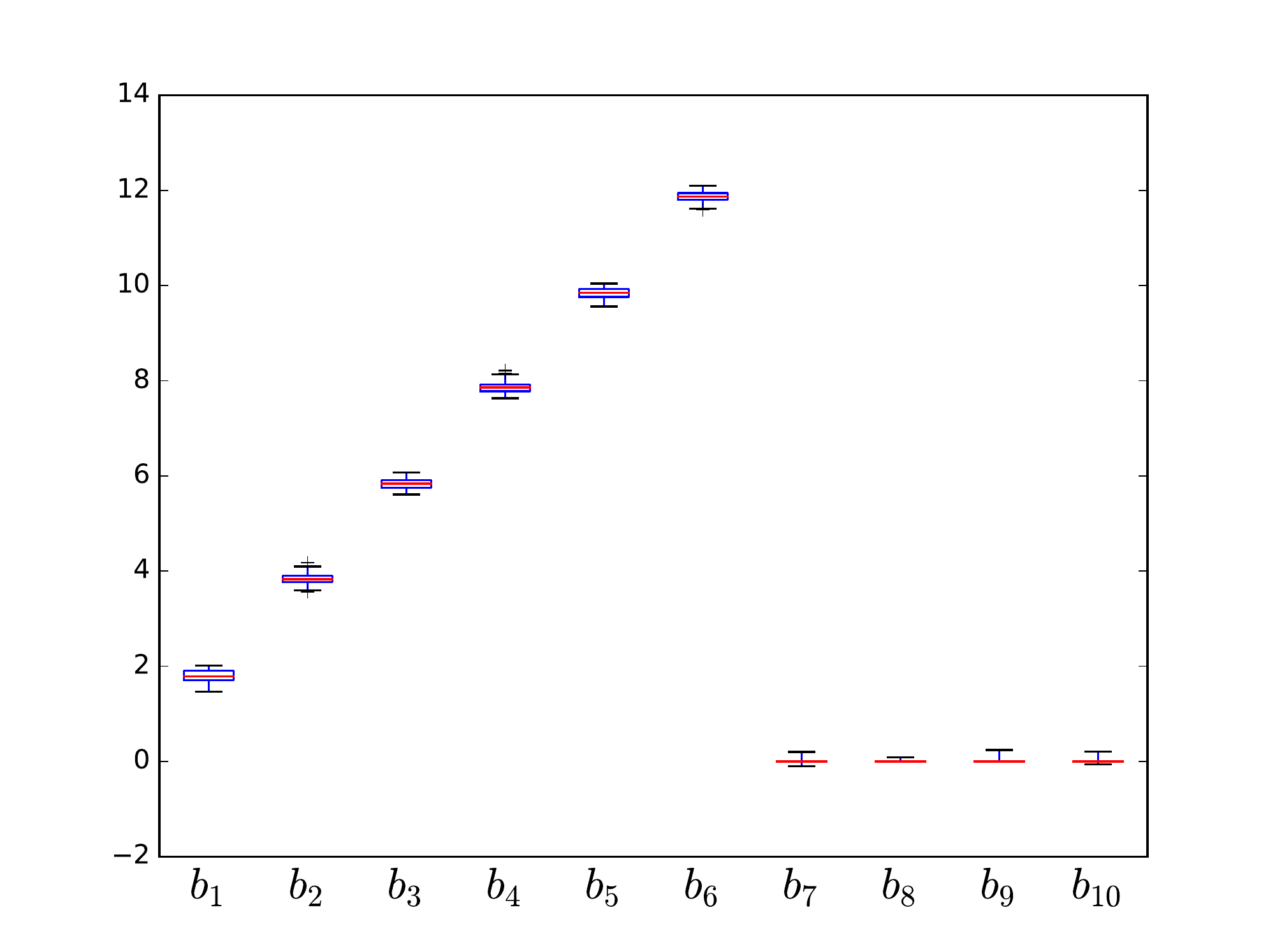}}
\subfloat[OLS estimates]
{\includegraphics[width=0.35\paperwidth]{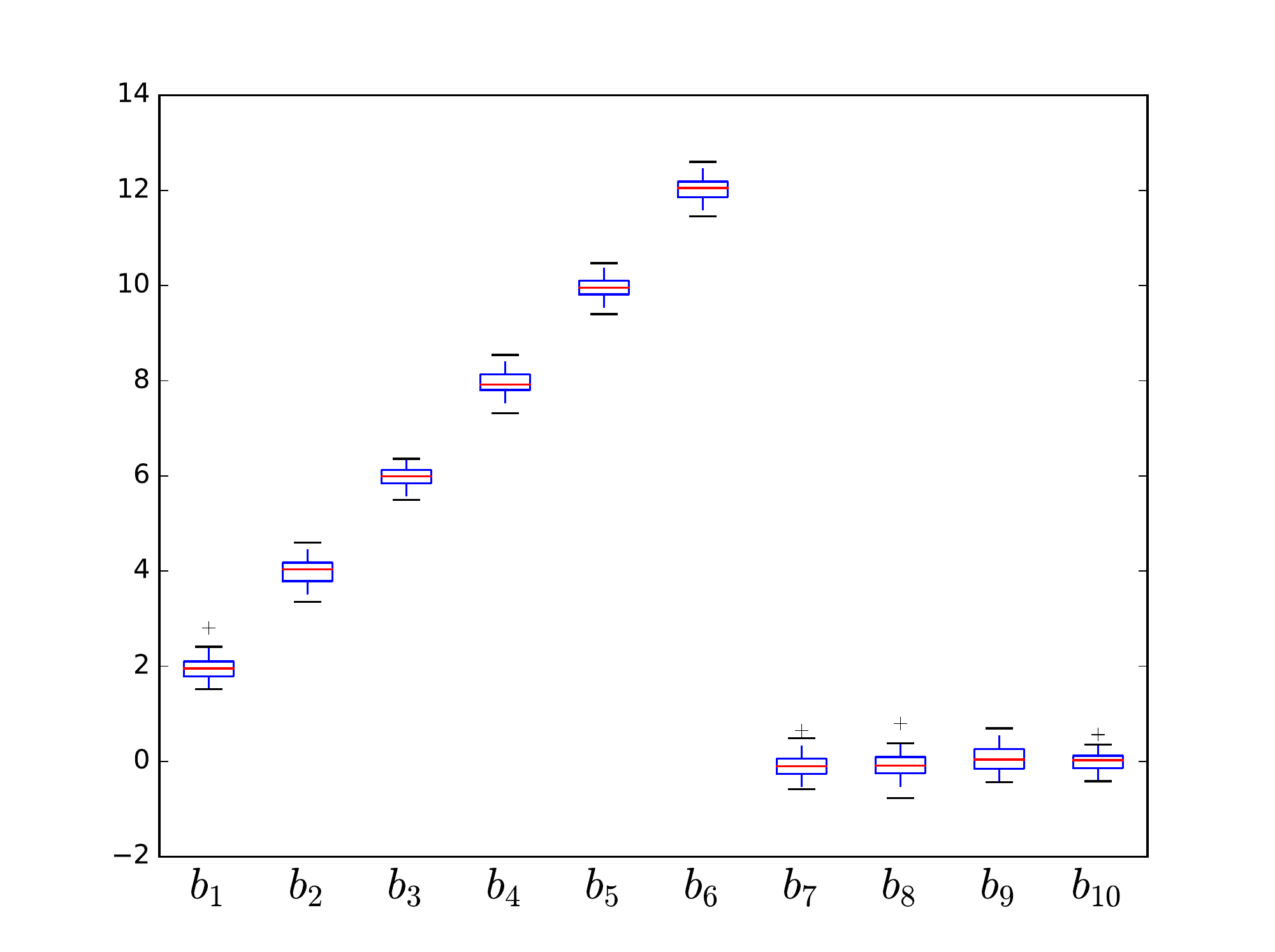}}

\subfloat[Histogram of $GR^{2}$]
{\includegraphics[width=0.6\paperwidth]{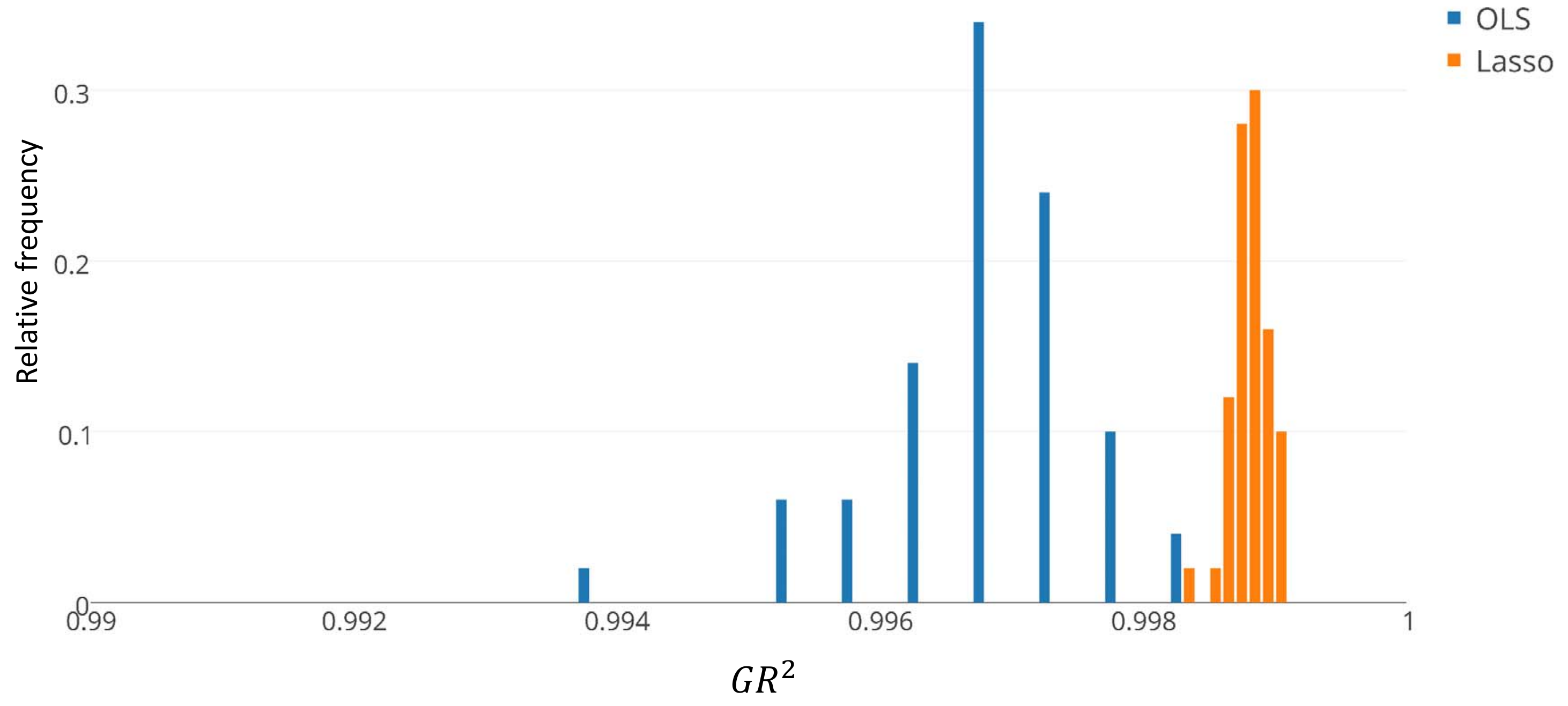}}

\caption{Boxplots of estimates and $GR^2$ for DGP $n=250,\thinspace p=200$ }
\end{figure}

\begin{figure}[ht]
\centering
\subfloat[Lasso estimates]
{\includegraphics[width=0.35\paperwidth]{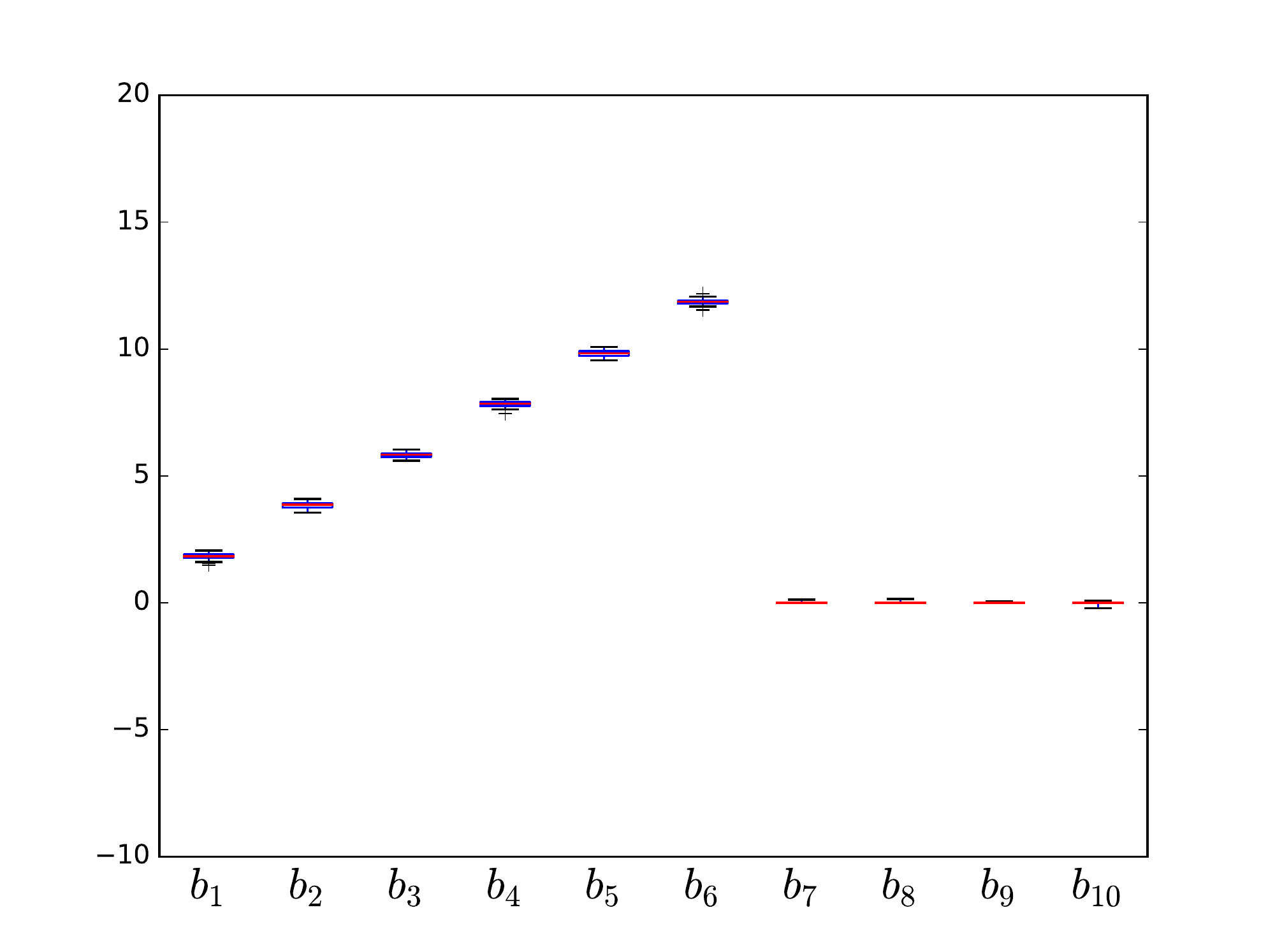}}
\subfloat[OLS estimates]
{\includegraphics[width=0.35\paperwidth]{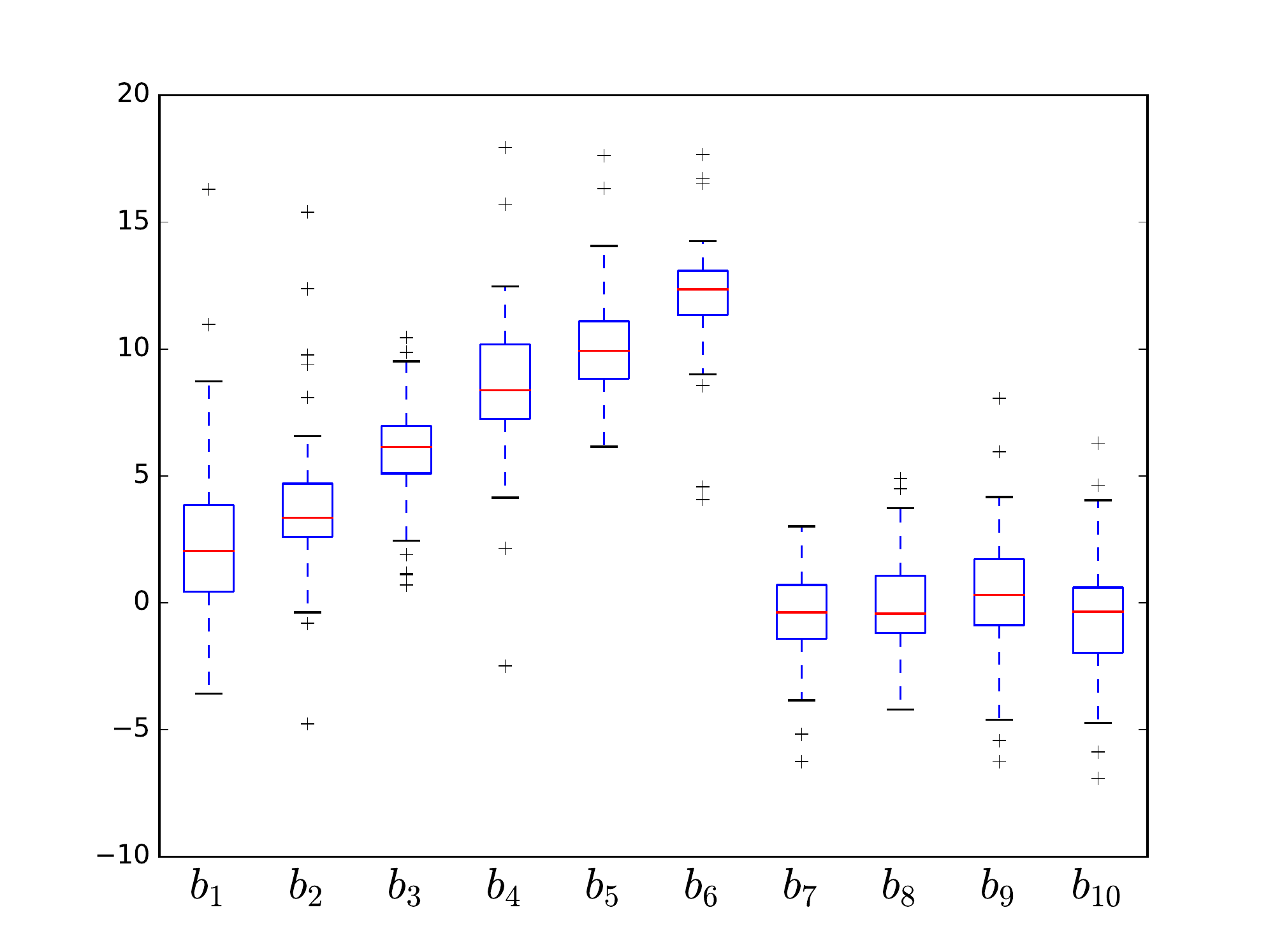}}

\subfloat[Histogram of $GR^{2}$]
{\includegraphics[width=0.6\paperwidth]{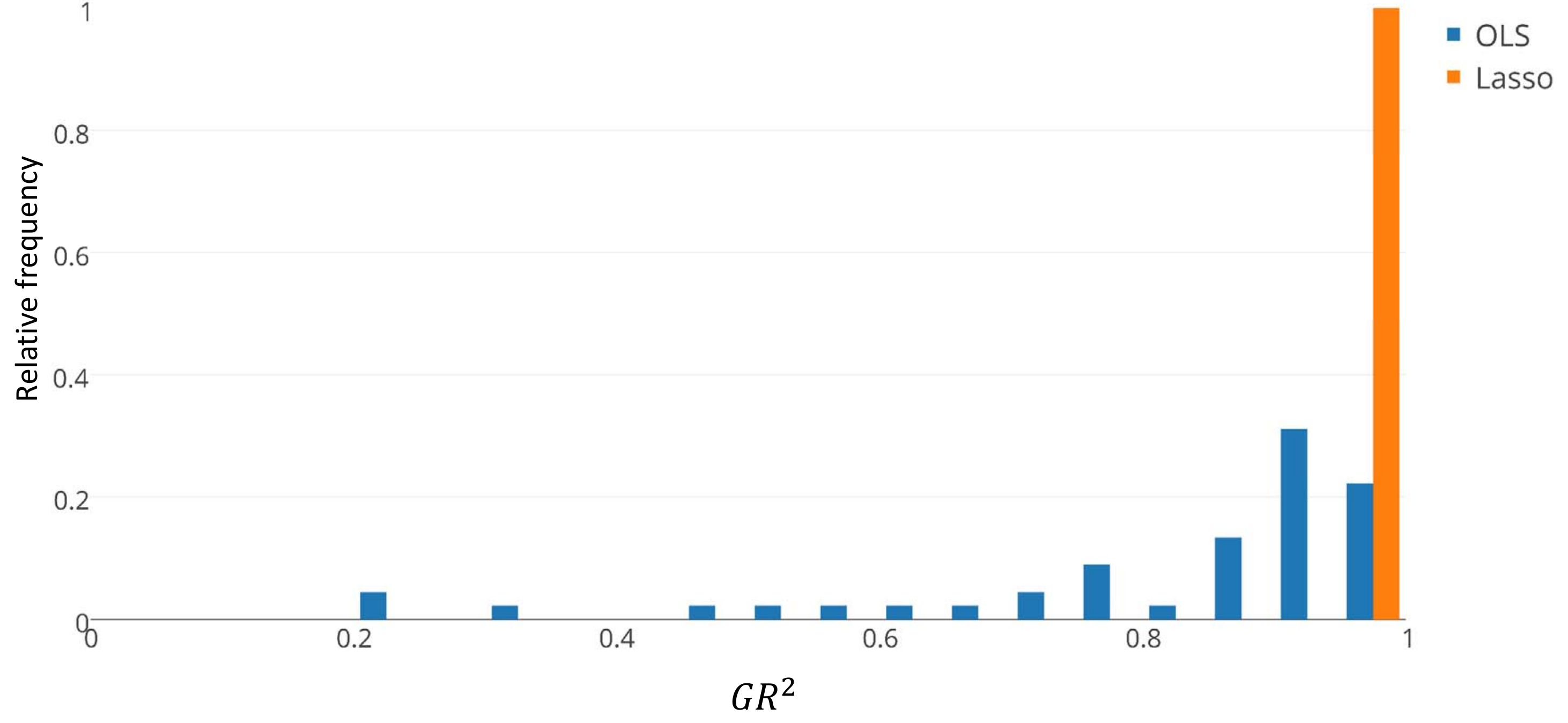}}

\caption{Boxplots of estimates and $GR^2$ for DGP $n=250,\thinspace p=250$ }
\end{figure}

\begin{figure}[ht]
\centering
\subfloat[Lasso estimates]
{\includegraphics[width=0.35\paperwidth]{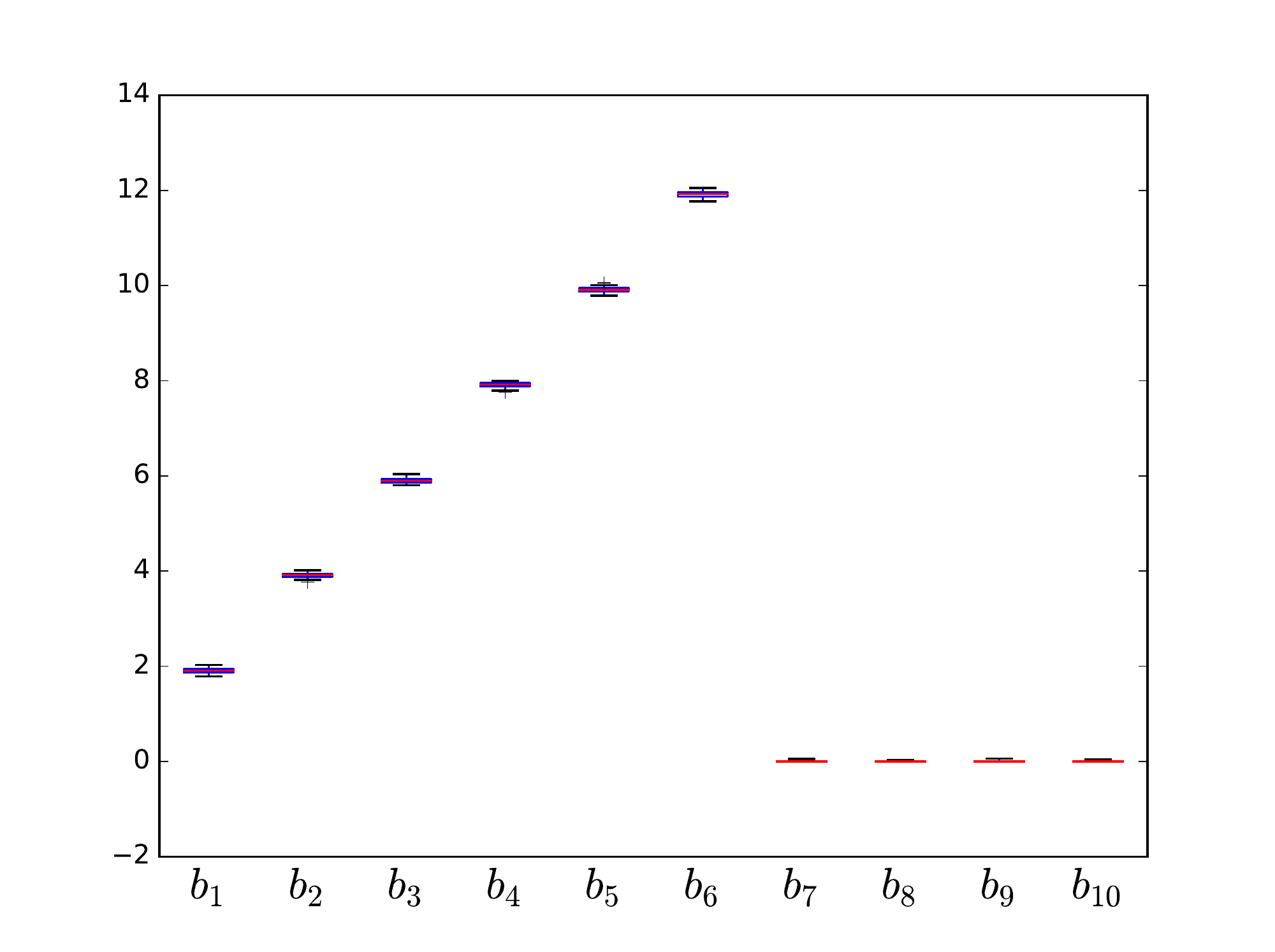}}
\subfloat[FSR estimates]
{\includegraphics[width=0.35\paperwidth]{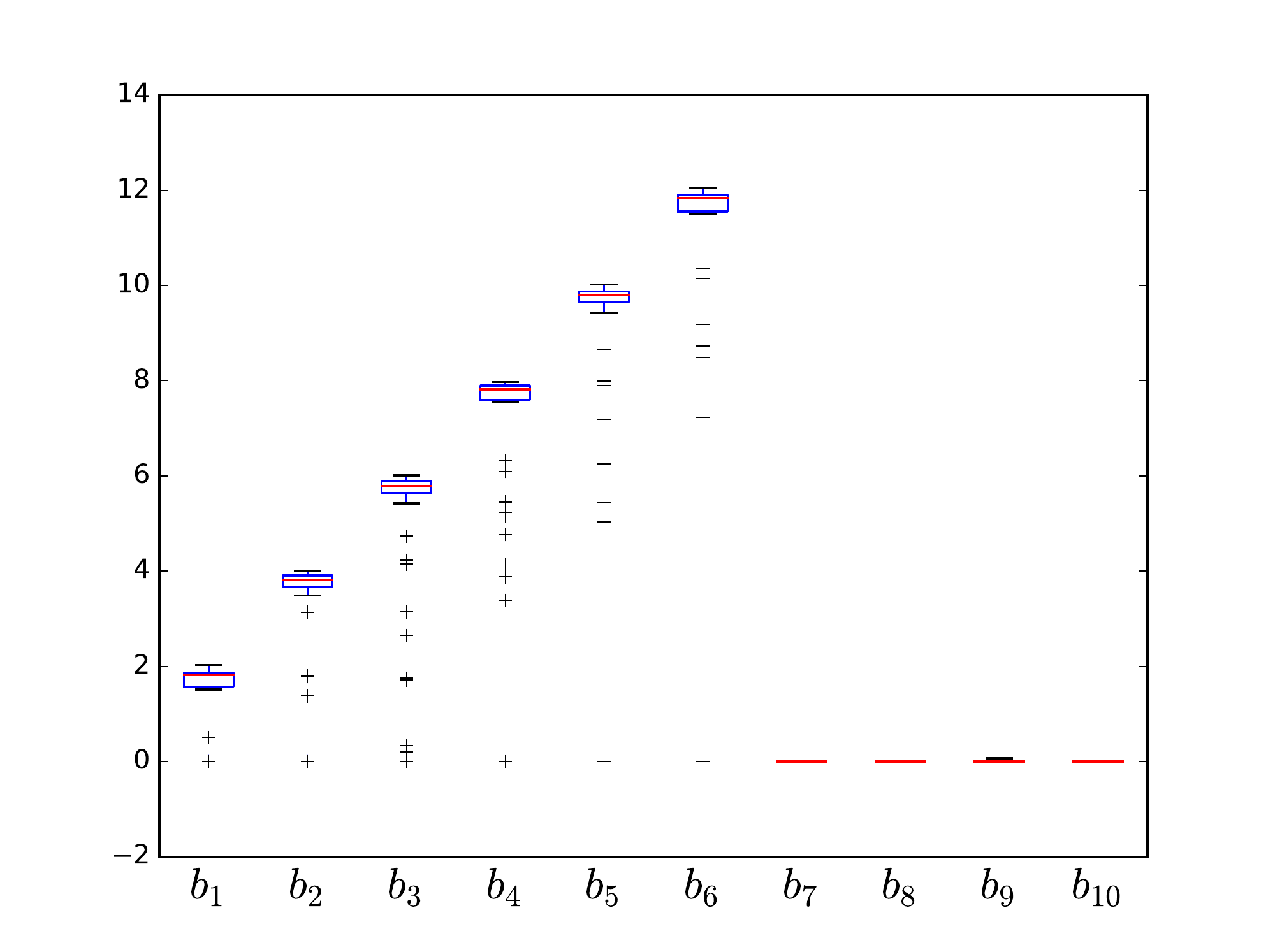}}

\subfloat[Histogram of $GR^{2}$]
{\includegraphics[width=0.6\paperwidth]{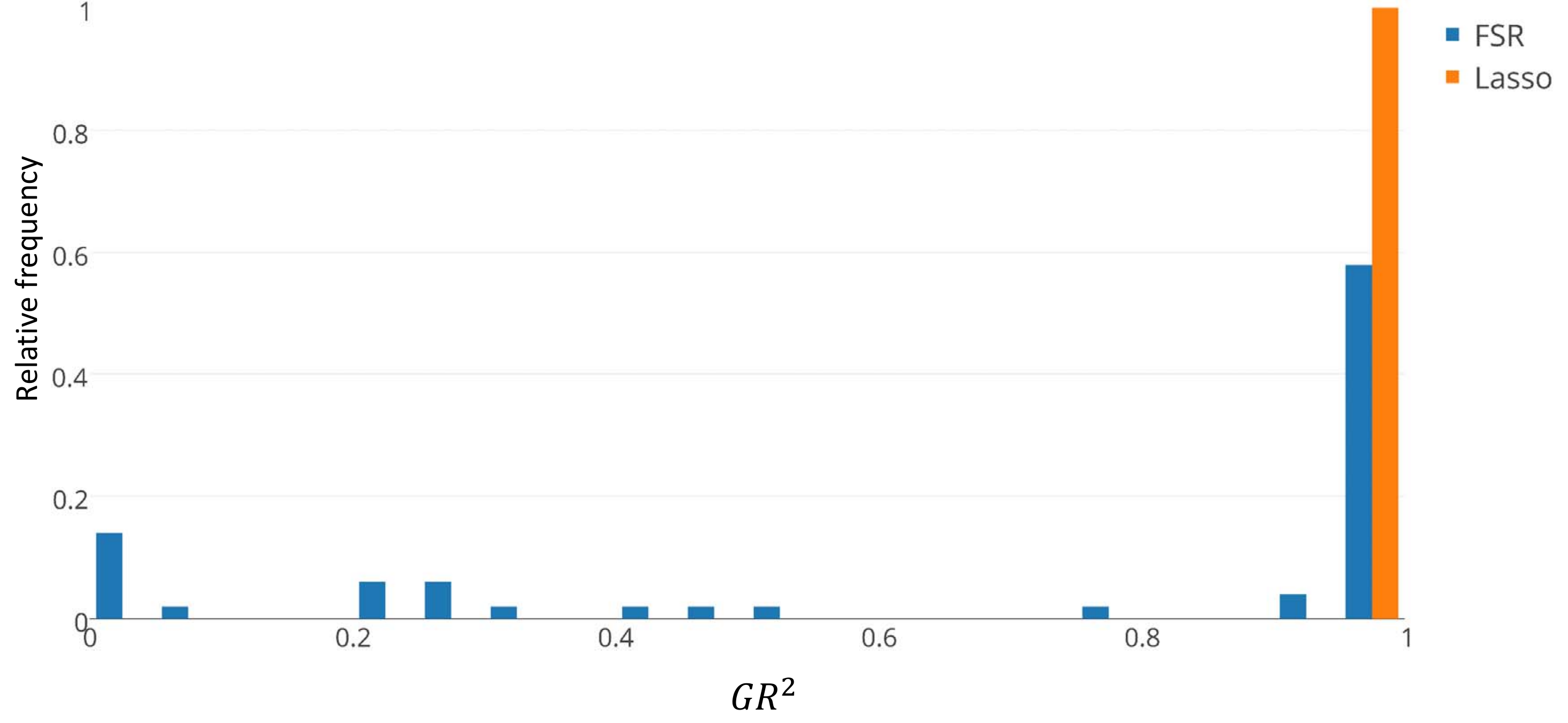}}

\caption{Boxplots of estimates and $GR^2$ for DGP $n=250,\thinspace p=300$ }
\end{figure}

\begin{figure}[ht]
\centering
\subfloat[Lasso estimates]
{\includegraphics[width=0.35\paperwidth]{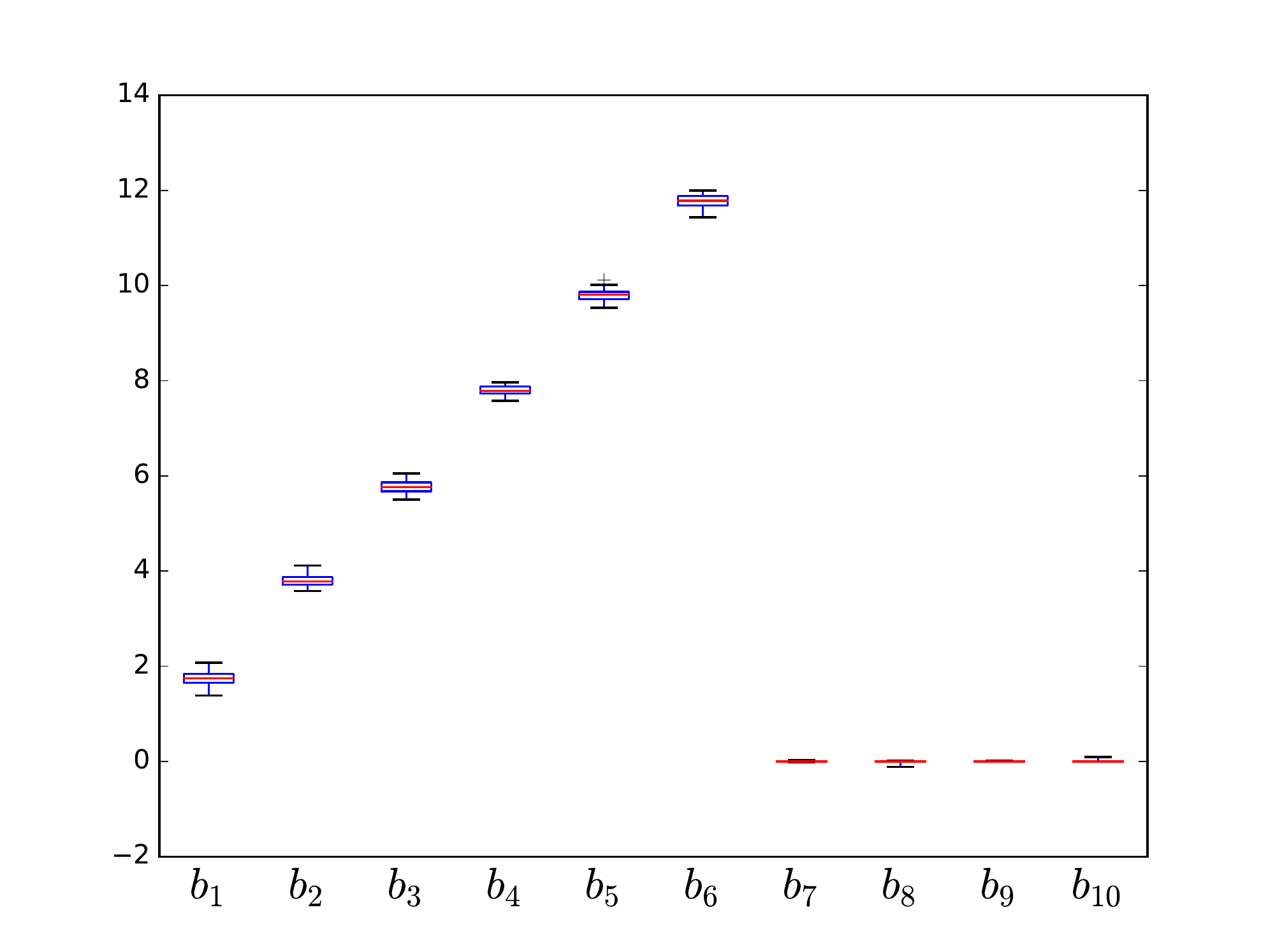}}
\subfloat[FSR estimates]
{\includegraphics[width=0.35\paperwidth]{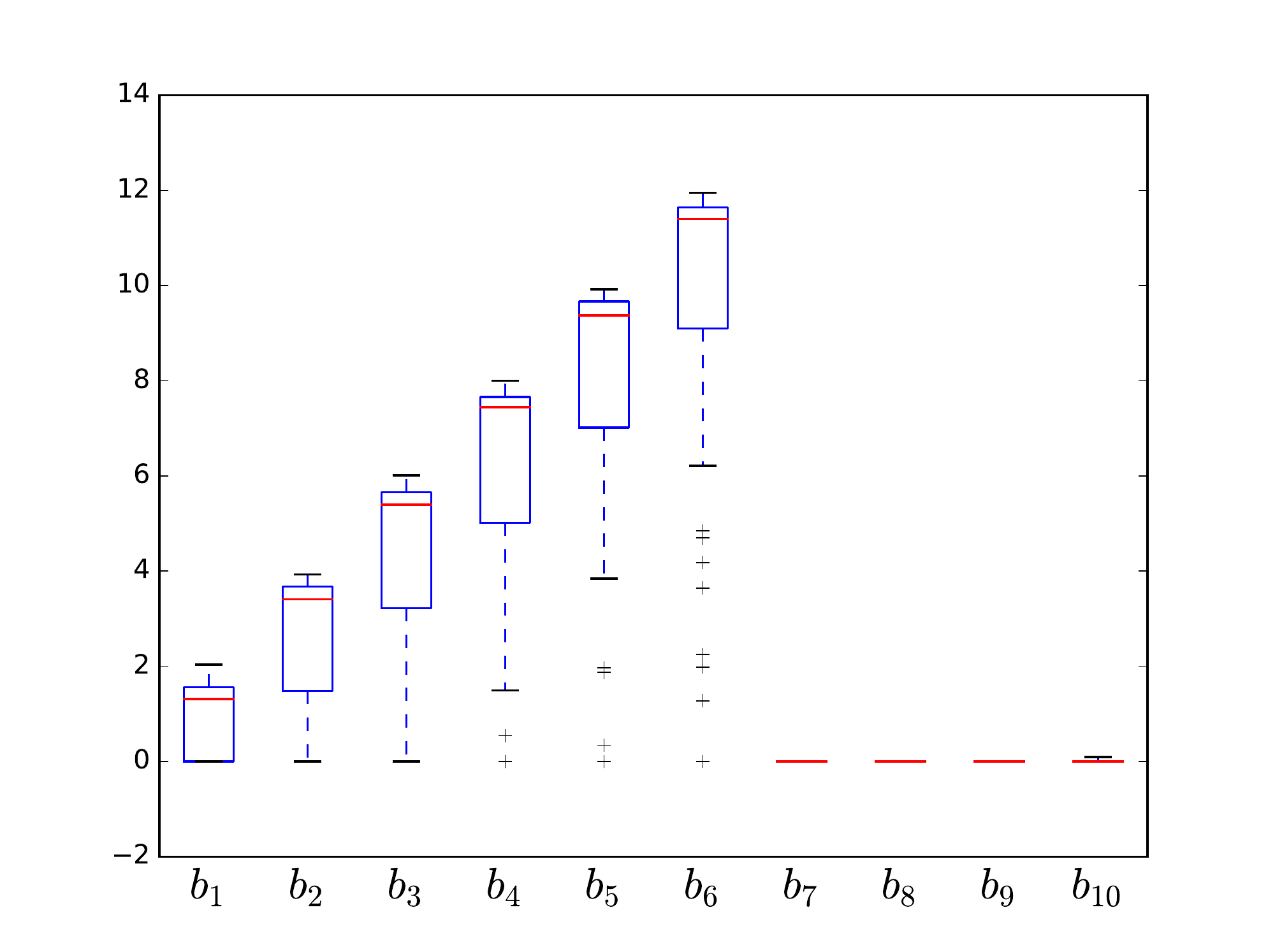}}

\subfloat[Histogram of $GR^{2}$]
{\includegraphics[width=0.6\paperwidth]{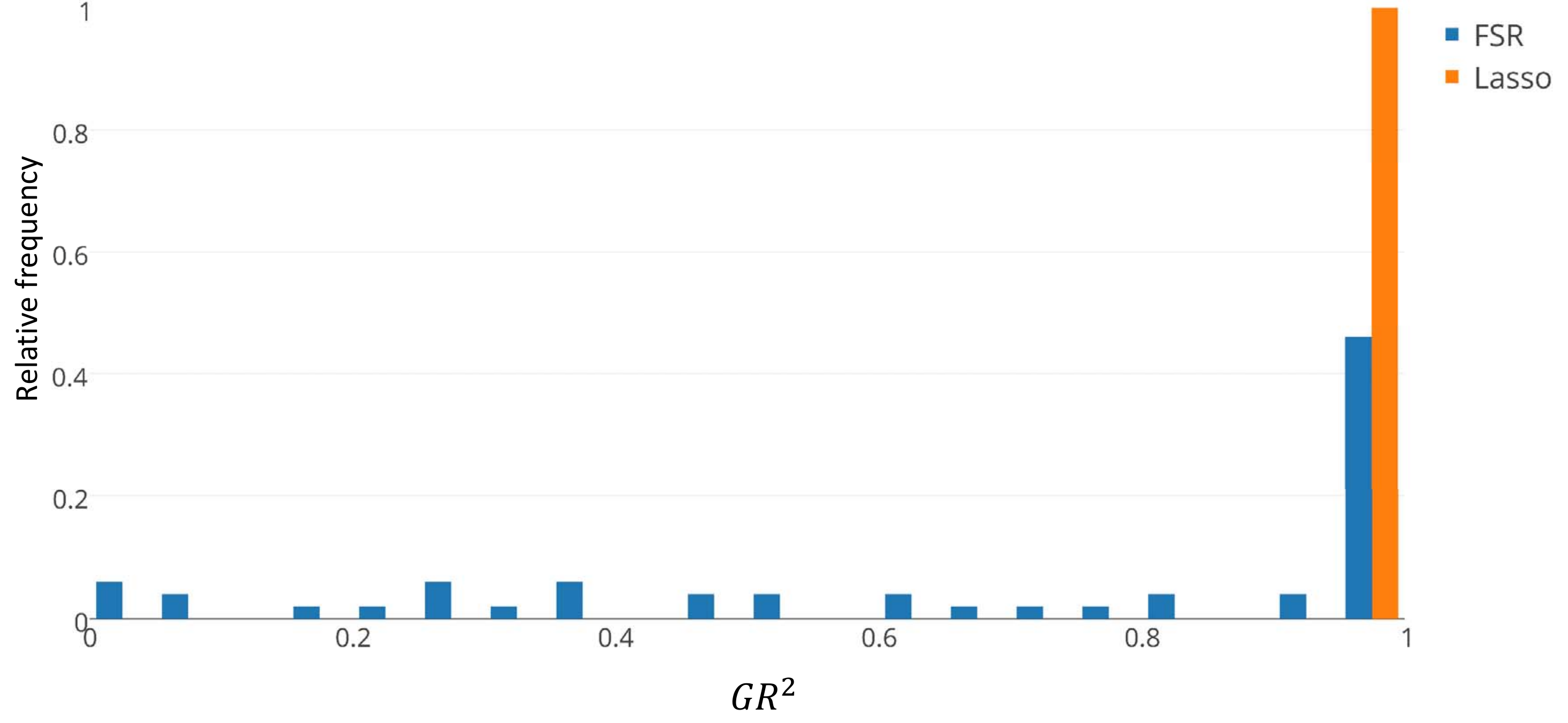}}

\caption{Boxplots of estimates and $GR^2$ for DGP $n=250,\thinspace p=500$ }
\end{figure}

\end{document}